\documentclass[sigconf]{acmart}

\AtBeginDocument{%
  \providecommand\BibTeX{{%
    \normalfont B\kern-0.5em{\scshape i\kern-0.25em b}\kern-0.8em\TeX}}}
    
\usepackage{algorithm}
\usepackage{algorithmic}

\usepackage{siunitx}
\usepackage{multirow}
\usepackage[T1]{fontenc}    % use 8-bit T1 fonts
\usepackage{hyperref}       % hyperlinks
\usepackage{url}            
\usepackage{booktabs}       % professional-quality tables
\usepackage{amsfonts}       % blackboard math symbols
\usepackage{nicefrac}       % compact symbols for 1/2, etc.
\usepackage{microtype}      % microtypography
\usepackage{lipsum}
\usepackage{subfigure}
\usepackage{graphicx}
\usepackage{amsmath,bm}
\usepackage{xcolor}         % \textcolor

\usepackage{algorithm}
\usepackage{algorithmic}

\usepackage{amsmath}

\copyrightyear{2021} 
\acmYear{2021} 
\setcopyright{rightsretained} 
\acmConference[WSDM '21]{Proceedings of the Fourteenth ACM International Conference on Web Search and Data Mining}{March 8--12, 2021}{Virtual Event, Israel}
\acmBooktitle{Proceedings of the Fourteenth ACM International Conference on Web Search and Data Mining (WSDM '21), March 8--12, 2021, Virtual Event, Israel}\acmDOI{10.1145/3437963.3441727}
\acmISBN{978-1-4503-8297-7/21/03}
\begin{document}
%\fancyhead{} 
\title{DeepLight: Deep Lightweight Feature Interactions for Accelerating CTR Predictions in Ad Serving}

\author{Wei Deng}
\authornote{Equal contribution}
\authornote{The work was done while Wei Deng was working as an intern at Yahoo Research; Deguang Kong is now affiliated with Google, Inc.}
\affiliation{%
  \institution{Purdue University}
  \city{West Lafayette, IN, USA}
  \country{weideng056@gmail.com}
}

\author{Junwei Pan}
\authornotemark[1]
\affiliation{%
  \institution{Yahoo Research}
  \city{Sunnyvale, CA, USA}
  \country{pandevirus@gmail.com}
}

\author{Tian Zhou}
\affiliation{%
  \institution{Yahoo Research}
  \city{Sunnyvale, CA, USA}
  \country{tian.zhou@verizonmedia.com}
}

\author{Deguang Kong}
\authornotemark[2]
\affiliation{%
  \institution{Yahoo Research}
  \city{Sunnyvale, CA, USA}
  \country{doogkong@gmail.com}
}

\author{Aaron Flores}
\affiliation{%
  \institution{Yahoo Research}
  \city{Sunnyvale, CA, USA}
  \country{aaron.flores@verizonmedia.com}
}

\author{Guang Lin}
\affiliation{%
  \institution{Purdue University}
  \city{West Lafayette, IN, USA}
  \country{guanglin@purdue.edu}
}

\begin{abstract}

Click-through rate (CTR) prediction is a crucial task in recommender system and online advertising. The embedding-based neural networks have been proposed to learn both explicit feature interactions through a shallow component and deep feature interactions by a deep neural network (DNN) component. These sophisticated models, however, slow down the prediction
inference by at least hundreds of times. To address the issue of significantly increased serving latency and high memory usage for real-time serving in production, this paper presents \emph{DeepLight}: a 
framework to accelerate the CTR predictions in three aspects: 1) accelerate the model inference via explicitly searching informative feature interactions in the shallow component; 2) prune redundant parameters at the inter-layer level in the DNN component; 3) prune the dense embedding vectors to make them sparse in the embedding matrix. By combining the above efforts, the proposed approach accelerates the model inference by 46X on Criteo dataset and 27X on Avazu dataset without any loss on the prediction accuracy. This paves the way for successfully deploying complicated embedding-based neural networks in real-world serving systems.

\end{abstract}




\maketitle

\section{Introduction}

%{ \color{red} new intro updated by dkong, please fill xxx if you can.}
% reference: https://dlsys.cs.washington.edu/pdf/lecture12.pdf

Online advertising has grown into a hundred-billion-dollar business since 2020, and the revenue has been increasing by more than 15\% per year for nine consecutive years~\cite{financial_report}. CTR prediction is critical in the online advertising industry, and the main goal is to deliver the right ads to the right users in the right context. Therefore, how to predict CTR accurately and efficiently has drawn the attention of both academic and industry communities.

Generalized linear models and factorization machine (FM) \citep{FM} have achieved great successes. However, they are limited in their prediction power due to the lack of mechanisms to learn deeper feature interactions. To tackle this issue, the embedding-based neural networks seek inspirations from computer vision and natural language processing and propose to include a shallow component to learn the informative low-order feature interactions and a DNN component for powerful high-order interaction modeling. In particular, Wide \& Deep~\cite{deepwide} proposed to train a joint network that combines a linear model and a DNN model, which, however, still requires feature engineering and is not end-to-end. DeepFM~\cite{deepfm} solved that problem by learning low-order feature interactions through the FM component instead of the linear model. Since then, various embedding-based neural networks have been proposed to improve the performance: Neural Factorization Machines (NFM)~\cite{NFM} uses a bilinear interaction pooling to connect the embedding vectors with the DNN component; Deep \& Cross Network (DCN)~\cite{deepcross} models cross features of bounded degrees in terms of layer depth; eXtreme Deep Factorization Machine (xDeepFM)~\cite{xdeepfm} incorporates a Compressed Interaction Network (CIN) \cite{deepcross} and a DNN to automatically learn high-order feature interactions in both explicit and implicit manners; AutoInt \cite{autoint} proposes to model high-order feature interactions using self-attentive neural networks with residual connections; AutoFIS \cite{autofis} proposes to identify important high-order feature interactions in the factorization machines for a more effective representation. Other extensions of embedding-based neural networks include \cite{PNN, deepcrossing, FLEN20, DIN}.

Despite the advances of these novel models applied in click prediction tasks in online advertising, the prediction is slowed down by hundreds of times compared to simple models such as logistic regression or factorization machine. This leads to unrealistic latency for the real-time ad serving system. One question that naturally follows is: are we able to serve the  high quality deep models with satisfactory model latency and resource consumption for real-time response in ad serving? 

Towards this goal, a practical solution needs to address
the following challenges (C1-C3). 

(C1) High quality: the served ``slim''
model is expected to be as accurate as the the original ``fat'' model.  

(C2) Low latency: the serving latency should be at very low level to maintain high QPS (Query per second) with few timeout.

(C3) Low consumption: memory costs should be low for pulling model checkpoint and storing them in memory in online ad serving. 

However, all the existing embedding-based neural networks, such as DeepFM, NFM, xDeepFM, and AutoInt, still focus on increasing the model complexity to achieve (C1) while sacrificing the performance in (C2) and (C3).  Whereas a few approaches, such as AutoCross \citep{autocross}, are proposed to improve the model efficiency, they didn't adopt the DNN framework and fail in achieving the state-of-the-art. To address these challenges all together, we propose an efficient model, so-called field-weighted embedding-based neural network (DeepFwFM) by improving the FM module via a field pair importance matrix, which is empirically as powerful as xDeepFM but becomes much more efficient. As shown in Figure \ref{sparse_r}, each component of DeepFwFM has an approximately sparse structure, which implies an advantage in structural pruning and potentially leads to an even more compact structure. By pruning the DNN component of DeepFwFM and further compressing the shallow component, the resulting deep lightweight structure, \emph{DeepLight}, greatly reduces the inference time and still maintains the model performance. By contrast, the other
structures may fail in either deep model accelerations \citep{NFM, deepcross, xdeepfm, autofis} or accurate predictions \citep{deepfm, deepwide, autoint}.

To the best of our knowledge, this is the \emph{first} paper that studies pruning embedding based DNN models for accelerating CTR predictions in ad serving. To summarize, the proposed field-weighted embedding-based neural network (DeepFwFM) has a great potential in fast and accurate inference. Compared to the existing embedding-based neural networks, the model has the following advantages:

$\bullet$ To address the challenge of (C1) \emph{high quality}, the proposed model exploits the field pair importance idea in FwFM \citep{fwfm,mt-fwfm} to improve the understanding of low-order feature interactions instead of exploring high-order feature interactions \citep{deepcross, xdeepfm}. Most notably, such an improvement achieves the state-of-the-art with \emph{a very low complexity} compared to the state-of-the-art xDeepFM model and still shows a great potential in deep model accelerations.

$\bullet$ To address the challenge of (C2) \emph{low latency}, the model can be pruned for further acceleration: 1) prune redundant parameters in the deep component to obtain the most accelerations; 2) remove the weak field pairs in FwFM to obtain additional significant accelerations. The resulting lightweight structure ends up with almost a hundred times of accelerations.

$\bullet$ To address the challenge of (C3) \emph{low consumption}, we can promote sparsity on the embedding vectors and preserve the most discriminant signals, which yields a substantial compression w.r.t. number of parameters.

By overcoming these challenges (C1-C3), the resulting sparse DeepFwFM, so-called \emph{DeepLight}, eventually achieves remarkable performance not only in predictions, but also in deep
model accelerations. It achieves \emph{46X speed-ups} on Criteo dataset and \emph{27X speed-ups} on
Avazu dataset without loss on AUC. To help reproduce the results and benefit the community, we have made code available at $\text{https://github.com/WayneDW/sDeepFwFM}$.

\section{Related Work}

CTR prediction is typically achieved by feeding a machine learning model with well-designed features \cite{Xinran_He, linear1}. There
have been extensive efforts on building features, such as textual
features \cite{Shaparenko}, click feedback features \cite{Graepel}, contextual features \cite{linear1}, and psychology features \cite{Taifeng}. To avoid the costly feature engineering, we need to build an end-to-end model by modeling feature interactions automatically. Given a dataset $\mathcal{D}=\{(y_i, \bm{x}_i)\}$, where $y_i$ is the label and $\bm{x}_i$ is a $m$-dimensional sparse feature vector, we can consider a degree-2 polynomial $\phi_{\text{Poly2}}$
\begin{equation}
    \label{logit}
    \min_{\bm{w}} \frac{\lambda}{2} \|\bm{w}\|_2^2 + \sum_{i=1}^{|\mathcal{D}|} \log(1+\exp(-y_i \phi_{\text{Poly2}}(\bm{w}, \bm{x_i}))),
\end{equation}
where $\bm{w}$ is the model parameter, $\lambda$ is the $L_2$ penalty, $|\mathcal{D}|$ is the number of data points, and 
\begin{equation}
\label{poly2}
        \phi_{\text{Poly2}}(\bm{w}, \bm{W})=w_0+\sum_{i=1}^m x_i w_i+\sum_{i=1}^m\sum_{j=i+1}^m x_i x_j W_{i,j}.
\end{equation}

Estimating the feature interaction matrix $\bm{W}$ in (\ref{poly2}) given insufficient data is not easy, FM \cite{FM} proposed a matrix decomposition method to learn the $k$-dimensional embedding vectors $\{\bm{e_i}\}_{i=1}^m$ through the inner product $\langle \bm{e}_i, \bm{e}_j\rangle$:
\begin{equation}
\label{fm}
        \phi_{\text{FM}}(\bm{w}, \bm{e})=w_0+\sum_{i=1}^m x_i w_i+\sum_{i=1}^m\sum_{j=i+1}^m x_i x_j \langle \bm{e}_i, \bm{e}_j\rangle.
\end{equation}

Notably, the embedding vectors $\{\bm{e}_i\}_{i=1}^m$ reduces the number of parameters from $O(m^2)$ to $O(mk)$. Although the above idea can be naturally extended to higher-order FMs, there is no efficient algorithm for training these models \citep{NIPS2016_6144}. As such, Deep \& Cross \citep{deepcross} tackles that problem by a series of cross operations:
\begin{equation}
    \bm{s}_l= \bm{s}_0 \bm{s}^T_{l-1} w_l+\bm{b}_l+ \bm{s}_{l-1},
\end{equation}
where $\bm{w}_l, \bm{b}_l, \bm{s}_l$ denote the weights, bias and output of the $l$-th layer of the DeepCross network, respectively. Such an improvement alleviates the computation problems in solving high-order FM models and becomes quite powerful in modeling low-order feature interactions. By further combining a DNN component, we can obtain xDeepFM\citep{xdeepfm}, which is the state-of-the-art model in CTR predictions.

\begin{figure*}[h!]
\centering
  \subfigure[DeepFM]{\includegraphics[scale=0.22]{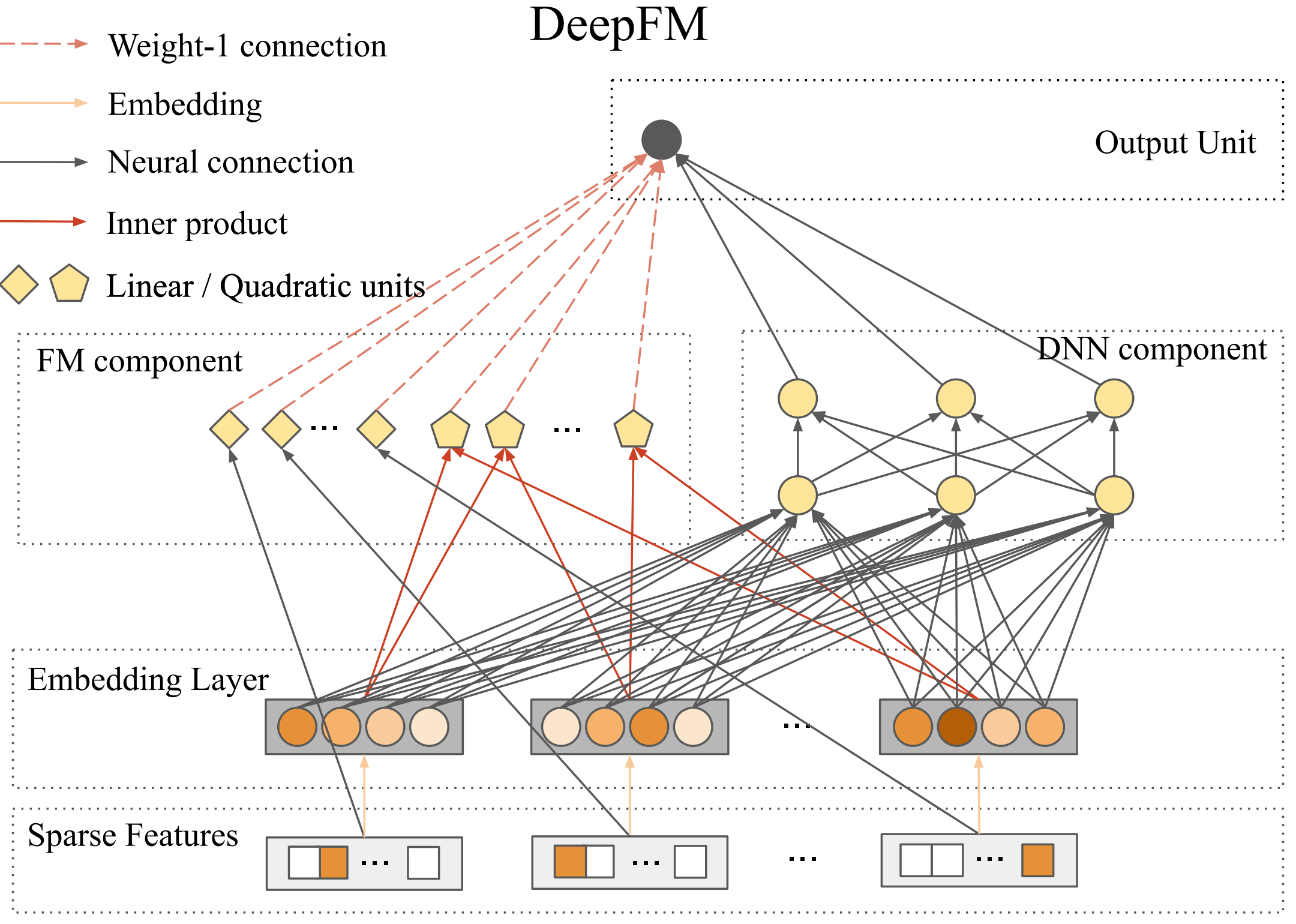}}\label{fig:2a}\qquad
  \subfigure[DeepFwFM (the proposed unpruned model)]{\includegraphics[scale=0.22]{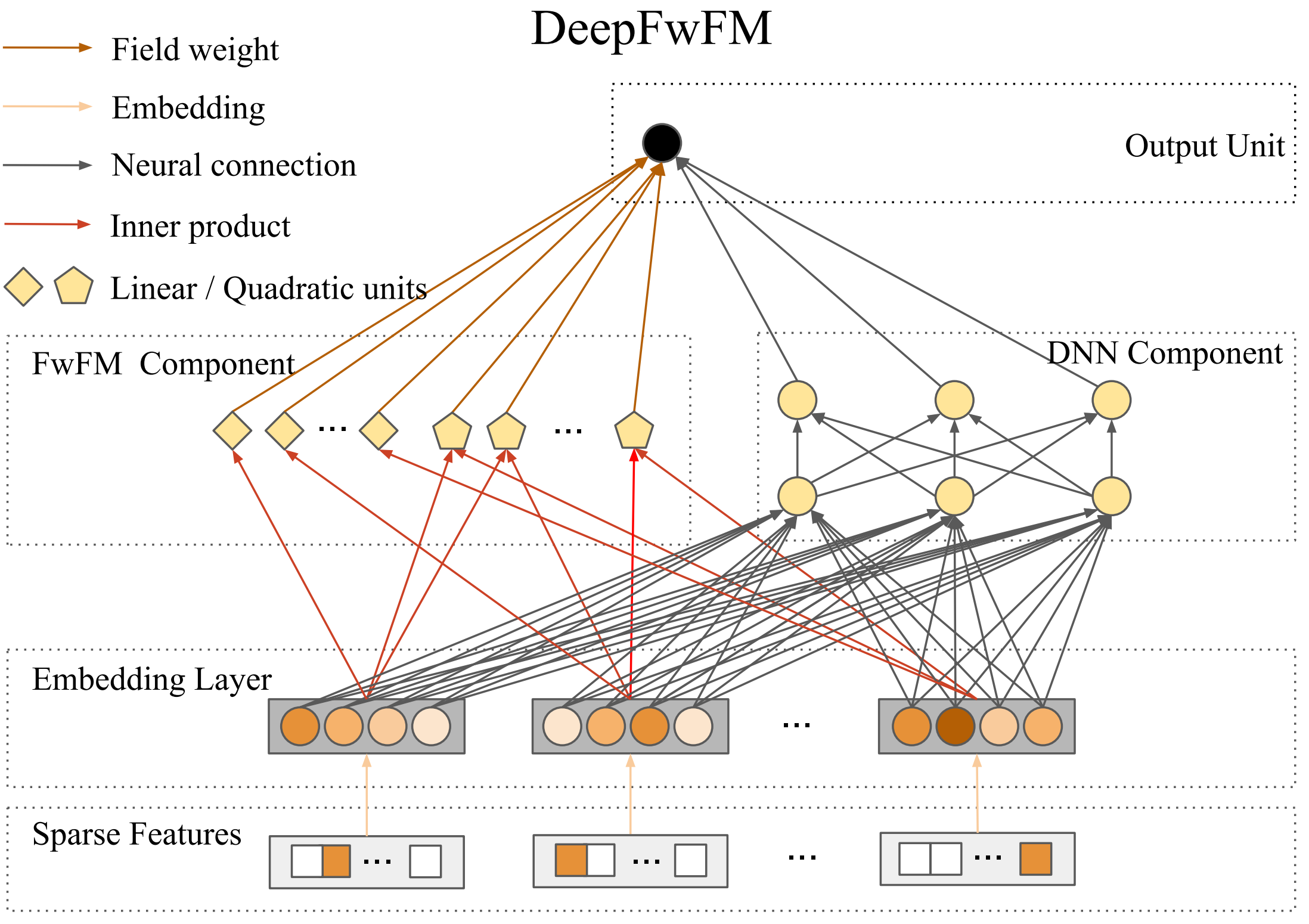}}\label{fig:2b}
  \caption{A model architecture comparison between DeepFM and proposed DeepFwFM. The inner products in the linear part of DeepFwFM are simplified. DNN component is generally built via the standard fully connected layers.}
  \label{deepFM}
%   \vspace{0.5em}
\end{figure*}

\section{DeepFwFM: An Efficient Model}

Considering the increasing latency concern of deploying complicated embedding-based neural networks in the real-world serving system, our goal is not only to build a model as accurate as xDeepFM~\cite{xdeepfm} but also to show the potential for significant deep model accelerations. Despite the powerful performance in modeling feature interactions, xDeepFM~\cite{xdeepfm} is known to be much more costly than DeepFM \citep{xdeepfm, autocross} and raises the risks in deep model accelerations. Such a problem motivates us to step back and rethink the framework of DeepFM. However, it is known that an ill-conditioned matrix suffers from a serious stability issue in matrix decomposition \citep{Michael_Jordan15, Li16} and affects the performance of DeepFM.

\subsection{Model architecture}

To solve the stability issue, we seek inspirations from the preconditioner for robust matrix decompositions and consider the FwFM model \citep{fwfm}, which includes the field pair importance to further improve the training of FM. In what follows, we propose the $\textbf{Deep}$ $\textbf{F}\text{ield-}$$\textbf{w}\text{eighted}$ $\textbf{F}\text{actorization }$ $\textbf{M}\text{achine}$ (DeepFwFM) by replacing $\phi_{\text{Poly2}}$ in (\ref{logit}) with $\phi_{\text{DeepFwFM}}$:
\begin{equation}
\begin{split}
    \label{deepfwfm_math}
        &\phi_{\text{DeepFwFM}}(\bm{w}, \bm{v}, \bm{e}, \mathbf{R})=\phi_{\text{Deep}}(\bm{w}, \bm{e}) + \phi_{\text{FwFM}}(\bm{v}, \bm{e}, \mathbf{R}),\\
\end{split}
\end{equation}
where $\bm{w}$ is the DNN parameter that contains the weights and bias, $\bm{v}\in \mathbb{R}^{n\times k}$, $\bm{e}$ denotes the set of embedding vectors, $\mathbf{R}\in \mathbb{R}^{n\times n}$ is a matrix to model field pair interaction strength, and $\phi_{\text{FwFM}}$ is an efficient model\citep{fwfm} in studying second-order feature interactions on multi-field categorical data such that
\begin{equation}
\label{fwfm__}
\small{
    \phi_{\text{FwFM}}(\bm{v}, \bm{e}, \mathbf{R})=w_0+\sum_{i=1}^m x_i \langle \bm{e}_i, \bm{v}_{F(i)} \rangle+\sum_{i=1}^m\sum_{j=i+1}^m x_i x_j \langle \bm{e}_{i}, \bm{e}_{j}\rangle R_{F(i), F(j)},}
\end{equation}
where $F(i)$ denotes the field of feature $i$, $\bm{v}_{F(i)}$ denotes the linear embedding vector for field $F(i)$, $R_{F(i), F(j)}$ denotes the field pair interaction weight between field pair $(F(i),F(j))$ and $\phi_{\text{Deep}}$ is a non-linear transformation of the embeddings through a Multilayer Perceptron to learn high-order feature interactions.

To summarize, our model has the following innovations:

1) DeepFwFM is much faster than xDeepFM in that we don't attempt to model high-order (3rd or more) feature interactions via the shallow component. Whereas xDeepFM contains a powerful compressed interaction network (CIN) to approximate any fixed-order of polynomials, the major drawback is that CIN has an even higher time complexity than the DNN component as discussed in \cite{xdeepfm}, resulting in expensive computations in large-scale ad systems;

2) DeepFwFM is more accurate than DeepFM because it overcomes the stability issue in matrix decomposition and leads to more accurate predictions \citep{ffm, fwfm} by considering the field pair importance. DeepFM models low-order feature interactions through the weight-1 connections between each hidden node in the FM component and the output node as shown in Fig.~\ref{deepFM}, which, however, fails to adapt to the local geometry and sacrifices the robustness.

\begin{figure*}[h!]
\centering
  \subfigure[Criteo: DNN component]{\includegraphics[scale=0.25]{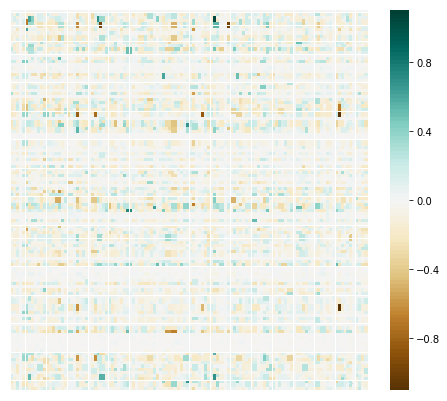}}\qquad\qquad
  \subfigure[Criteo: Field matrix $R$]{\includegraphics[scale=0.25]{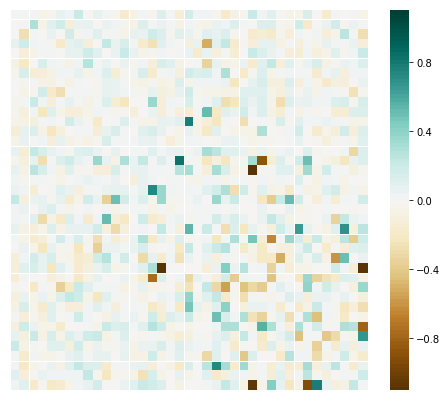}}\qquad\qquad
  \subfigure[Criteo: Embedding vectors]{\includegraphics[scale=0.25]{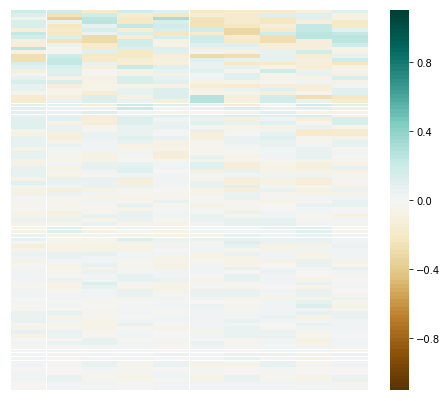}}
  \subfigure[Avazu: DNN component]{\includegraphics[scale=0.25]{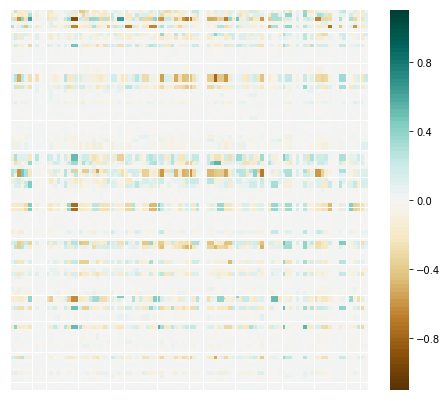}}\qquad\qquad
  \subfigure[Avazu: Field matrix $R$]{\includegraphics[scale=0.25]{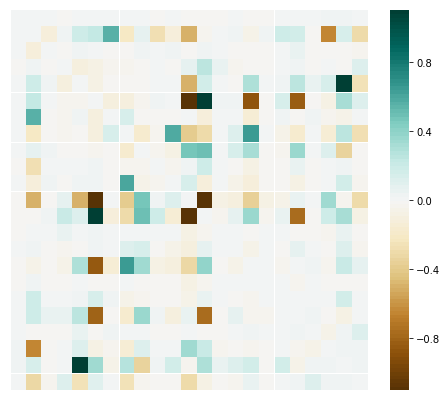}}\qquad\qquad
  \subfigure[Avazu: Embedding vectors]{\includegraphics[scale=0.25]{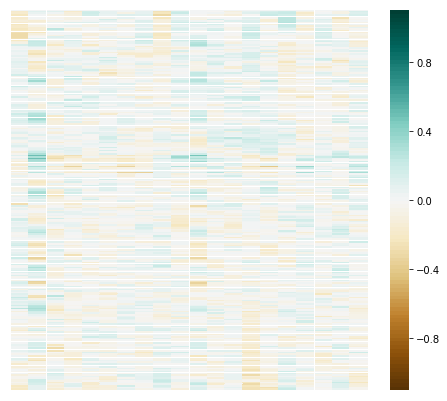}}
  \caption{Weight demonstration of DeepFwFM model on Criteo and Avazu datasets. In particular for the DNN component and embedding vectors, we only choose a representative part to present due to their complexity. Moreover, we apply the magnitude-based max pooling operation to large matrices for illustration purposes.}
  \label{sparse_r}
  \vspace{1em}
\end{figure*}

\subsection{Benefits of computation efficiency}
\label{complexity}

To demonstrate the benefits of our model, we perform quantitative analysis for computational complexity of DeepFwFM, compared to DeepFM and xDeepFM. 

\paragraph{Computational complexity} Given the embedding size $k$, the number of layers $l$, and the number of nodes in each layer $h$, the embedding layer only has $n$ lookups and leads to
little computational cost. The number of floating point operations (FLOPs) of the DNN component and the FwFM component is $\mathcal{O}(lh^2+nkh)$ and $\mathcal{O}(n^2k)$, respectively. Similarly, we can derive the computational complexity of DeepFM \citep{deepfm}
 and xDeepFM \citep{xdeepfm}. We summarize the results in Table.\ref{com_complexity} and observe that a $l$-layer CIN in xDeepFM takes $\mathcal{O}(nkh_c^2l)$ operations \cite{xdeepfm}, which is \emph{much more than the DNN component}. By contrast, the FM component and FwFM component is much faster than the DNN component. For example, considering standard parameter settings in Table.\ref{com_complexity}, we see that theoretically DeepFwFM is as efficient as DeepFM and is \emph{18X faster} than xDeepFM.

Despite the initial progresses on accelerating CTR predictions, DeepFwFM still fails to reduce the latency to a good level (e.g. 10 ms for each bid request). In fact, DeepFwFM can be \emph{hundreds-of-times slower} due to the inclusion of the DNN component, which significantly slows down the online CTR predictions.

\begin{table}
 \caption{Computational complexity for three standard deep models. For example, a popular choice in Criteo dataset is to set $l=3$, $h=400$, $n=39$, $k=10$, $h_c=100$. As a result, the computations for DeepFM, xDeepFM and DeepFwFM are of order 0.64M, \emph{12M} and 0.65M, respectively.}
  \centering
  \begin{tabular}{lccccr}
    \toprule
    % \multicolumn{4}{c}{Criteo Dataset}                   \\
    % \midrule
    Data     & shallow component & DNN component    \\
    \midrule
    DeepFM &  $\mathcal{O}(nk)$ & $\mathcal{O}(lh^2+nkh)$  \\
    xDeepFM & $\mathcal{O}(nkh_c^2l)$ & $\mathcal{O}(lh^2+nkh)$ \\
    DeepFwM & $\mathcal{O}(n^2k)$ & $\mathcal{O}(lh^2+nkh)$  \\
    \bottomrule
  \end{tabular}
  \label{com_complexity}
%   \vspace{-1em}
\end{table}

\section{DeepLight: A Lightweight model via Structural pruning}

The DNN component is undoubtedly the main reason that causes the high latency issue and fails to meet the online requirement. Therefore, a proper acceleration method is on great demand to speed up the predictions.

\subsection{Why structural pruning?}

Deep model acceleration has achieved great popularity in computer vision and it consists of three main methods: structural pruning \cite{han2015learning, li17}, knowledge distillation \cite{distill}, and quantization \cite{weight_quant1}, among which structural pruning methods have received wide attentions \cite{han2015learning, hansong16, deng2019, strucprunining, deguang, ye_icml, ye_nips} due to their remarkable performance. Moreover, each component of DeepFwFM, such as the DNN component, the field pair interaction matrix $R$ and the embedding vectors, possesses the highly-sparse structure as shown in Figure.\ref{sparse_r}. This motivates us to consider structural pruning to accelerate both the shallow component and the DNN component.

For the other choices, quantization ~\cite{hansong16} adopts the effective fixed point precision in inference time. However, it doesn't fully utilize the sparse structure of each component of DeepFwFM and even damages the precision of large coefficients. While knowledge distillation technique~\cite{distill} trains small networks (i.e. student model) to mimic larger ones (i.e., teacher model), it suffers from the following issues: %of inconsistency between teacher and student models including: 
a) the student model may have a limited capacity to learn from the teacher; 
% b) the teacher's prediction may not be available prior to student for online training for ad prediction; 
b) where there is a performance deterioration issue, it's hard to figure out whether it's due to the teacher model or due to the teaching procedure.

That's the reason why we adopt \emph{structure pruning} for CTR prediciton models in this work.

\subsection{How to do structural pruning?}

{\bf Design} In this section, we show how to do structural pruning for DeepFwFM. Instead of simply applying \emph{pruning} technique in ad prediction task with a uniform sparse rate,  we propose to prune the following three components (in the context of DeepFwFM model) that are particularly designed to adapt for ad prediction task (given the existing feature embedding and field relevance in shallow and deep components): 

$\bullet$ Prune the weights (excluding bias) of the DNN component to remove the neural connections;

$\bullet$ Prune the field pair interaction matrix $\bm{R}$ to remove redundant field interactions;

$\bullet$ Prune the elements in the embedding vectors, leading to sparse embedding vectors.

\begin{figure}[h!]
\centering
  \includegraphics[scale=0.22]{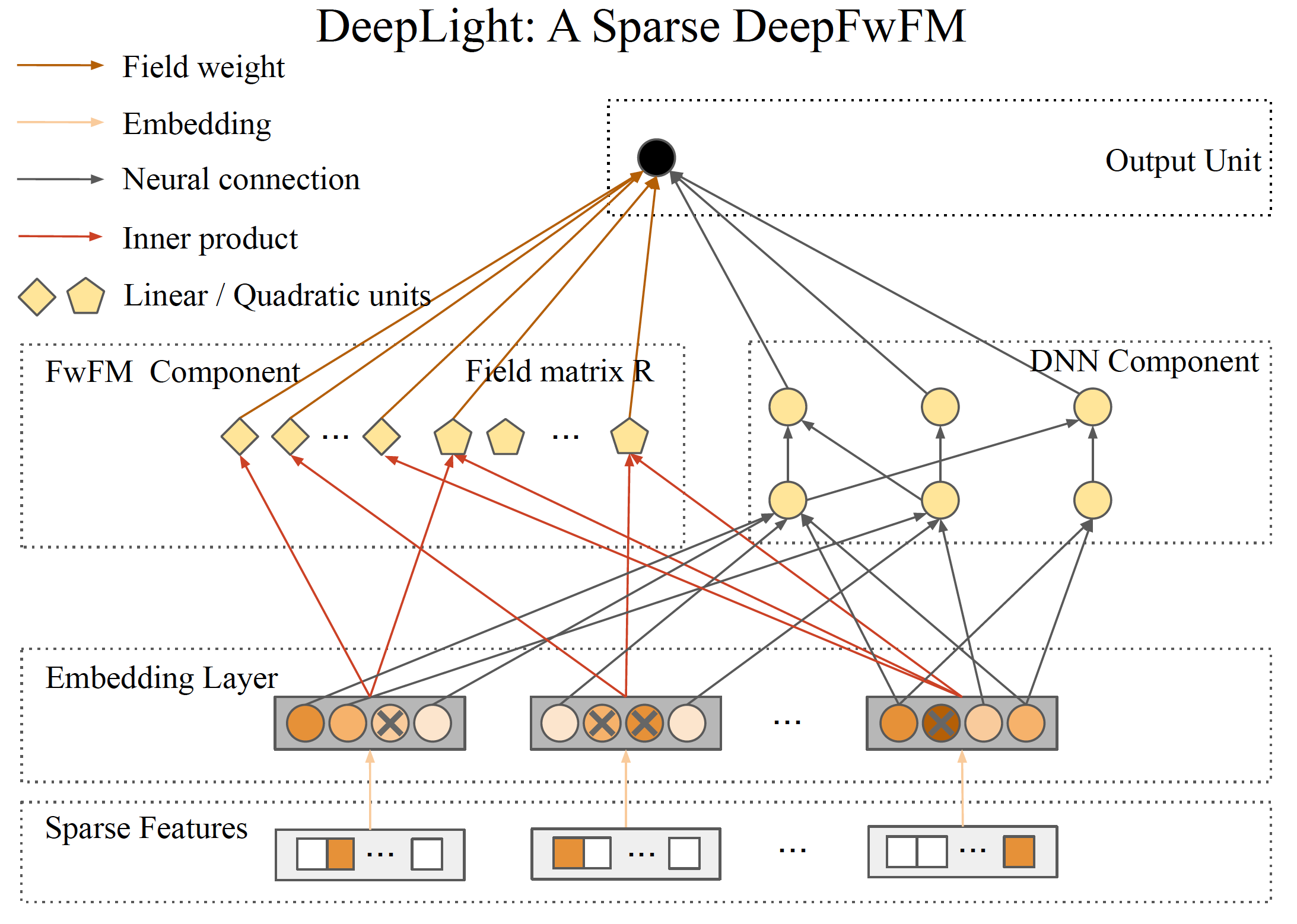}
  \caption{DeepLight: A sparse architecture of DeepFwFM. The inner products in the linear part are simplified.}
  \label{sparse_deepFwFM}
%   \vspace{2em}
\end{figure}

Combining the above efforts, we obtain the desired DeepLight model, as shown in Fig. \ref{sparse_deepFwFM}. DeepLight is a sparse DeepFwFM that provides a holistic approach to accelerate inference by  modifying the trained architecture at inference time.

{\bf Benefit} As is evident in empirical study, DeepLight enjoys the following properties

$\bullet$ The sparse DNN component has much less computation complexity, compared with the original one. It leads to the most accelerations in computation;

$\bullet$ The sparse field pair interaction matrix $\bm{R}$ further achieves significant accelerations in the FwFM component. In addition, pruning $\bm{R}$ is actually doing feature selection, or more accurately, \emph{field pair selection}. Once a field pair interaction weight $R_{F, F'}$ is pruned, all feature pairs from field pair $(F, F')$ are dropped. AutoFIS~\cite{autofis} also achieves a similar kind of field pair selection;

$\bullet$ The sparse embedding greatly reduces the memory usage, since the feature embeddings dominate the number of parameters in deep learning models for click prediction.

{\bf Implementation} Selecting a good sparse network from an over-parameterized model is NP-hard, and no optimal algorithms are guaranteed to solve it. There are a lot of empirical studies existed in the area of structural pruning \cite{han2015learning, hansong16, frankle2018lottery, deng2019}, including weight pruning, neuron pruning, or pruning of other units. Considering the fact that our model only contains a standard FwFM component and a vanilla DNN component, we conduct weight pruning and seek to achieve high sparsity and accelerations without calling specifically dedicated libraries.

Regarding the implementations, we adopt a standard algorithm used in \citep{deng2019, frankle2018lottery} to study the structural pruning of the mainstream CTR-prediction models and compare them with our proposed sparse DeepFwFM, namely DeepLight. As in most of the pruning works, we put the community wisdom in our algorithm by adopting the $L_2$ penalty \citep{han2015learning, hansong16}. Now, we present the main algorithm in Alg.\ref{prune}. We first train the model a few epochs to provide a good initialization and then conduct pruning to remove the redundant weights with the lowest magnitude. After each pruning, we retrain the model to fine-tune the network so that mistakenly pruned weights has a potential to become active again. We keep repeating this pruning procedure and set adaptive sparse rates such that the rate increases faster in the early phase when the network is stable and slower in the late phase when the network becomes sensitive. The algorithm behaves similar to the greedy algorithms based on the weak sub-modular optimization \cite{Abhimanyu11}, but it also injects some additional uncertainty during the pruning process, which potentially avoids damaging good weights.

\begin{algorithm}
\caption{Structural pruning for a target model, the target sparse rate $S\%=99\%$ means 99\% of the parameters are pruned.}\label{prune}
\begin{algorithmic}[1]
\STATE{\textbf{Input} Set the target sparse rate ${S}$, damping ratios $\mathbb{D}$ and $\mathbb{f}$.}
\STATE{\textbf{Warm up} Initialize a neural network by training $i$ epochs.}
\STATE{\textbf{Pruning}

For $k=1, 2, ...$ do

\ \ \ \ Train the network for one iteration.

\ \ \ \ Enumerate the candidate component $X$ in a model

\ \ \ \ \ \ \ \ Update the current sparse rate $s_X\leftarrow {S_X}(1-\mathbb{D}^{k/{\mathbb{f}}})$.

\ \ \ \ \ \ \ \ Prune the bottom-$s_X\%$ lowest magnitude weights.}

\STATE{\textbf{Online Prediction}

Transform the sparse model to efficient structure.}
\end{algorithmic}
% \vspace{-0.1em}
\end{algorithm}

\subsection{Reduction of computational complexity} 

The DNN component is the bottleneck that causes the high inference time. After the pruning of the DNN component, the FwFM component becomes the limitations, which requires further pruning on the field matrix $\bm{R}$. The pruning of the embedding layer has no significant speed-ups on the computations. 

With a medium $S_{\text{dnn}}\%$ sparsity on the weights in the DNN component (excluding the bias), the corresponding speed-ups can be close to the ideal $1/(1-S_{\text{dnn}}\%)$ times. However, when the sparsity $S_{\text{dnn}}\%$ is higher than 95\%, we may not achieve the ideal rate because of the computations in the biases and the overhead of sparse structures, such as the compressed row storage (CRS). As to the pruning of the field matrix $\bm{R}$, the speed-ups becomes more significant as we increase the sparsity $S_{\text{dnn}}\%$ in the DNN component.

\begin{table*}
 \caption{Model comparison on the Criteo and Avazu datasets.}
  \centering
%   \small
  \begin{tabular}{lccrc|ccrc}
    \toprule
    \multicolumn{5}{c|}{Criteo} & \multicolumn{4}{c}{Avazu} \\  
    \midrule
    \multirow{2}{*}{Models}    & \multicolumn{2}{c}{Test} & \multirow{2}{*}{\# Parameters} & \multirow{2}{*}{Latency ($ms$)}  & \multicolumn{2}{c}{Test} & \multirow{2}{*}{\# Parameters} & \multirow{2}{*}{Latency ($ms$)}  \\
      & LogLoss   & AUC &  &  & LogLoss & AUC & & \\
    \midrule
    LR      &   0.4615 & 0.7881 & 1,326,056 & 0.001 & 0.3904 & 0.7617 & 1,544,393 & 0.001 \\
    FM        & 0.4565 & 0.7949  & 14,586,606 & 0.005 &  0.3816 & 0.7782 & 32,432,233 & 0.009 \\
     FwFM        & 0.4466 & 0.8049  & 13,261,682 & 0.145 &  0.3764 & 0.7866  & 30,888,853 & 0.105 \\
    \midrule
    DeepFM    & 0.4495 & 0.8036 & 15,064,206 & 4.181 & 0.3780 & 0.7852 & 32,751,433 & 2.719    \\
    NFM        &  0.4497 & 0.8030  & 15,204,206 & 4.091 & 0.3777 & 0.7854 & 32,689,033 &  2.704 \\
    xDeepFM      & 0.4420 & 0.8102 & 15,508,958 & 40.85 &  0.3749 & 0.7894 & 32,927,058 & 7.129 \\
    \midrule
    DeepFwFM    & 0.4403 & 0.8116 & 13,739,321 & 4.271 & 0.3751 & 0.7893  & 31,208,053 & 2.824 \\
    \bottomrule
  \end{tabular}
  \vspace{1em}
  \label{criteo_data}
\end{table*}

\subsection{Reduction of space complexity} 

Pruning in the embedding layer also dramatically reduces the number of parameters in DeepFwFM and therefore saves lots of memory. In the embedding layer, a $S_{\text{emb}}\%$ sparsity reduces the number of parameters from $mk$ to $(1-S_{\text{emb}}\%)mk$. While in the DNN component, the number of weights (excluding the bias) can be reduced from $\mathcal{O}(nkh + lh^2)$ to $\mathcal{O}\left((nhk+lh^2)(1-S_{\text{dnn}}\%)\right)$ by storing the sparse weight matrix through the CRS. Similarly, a $S_{\text{R}}\%$ sparsity on the field matrix $\bm{R}$ reduces the parameters proportionally. Since the parameters in the embedding vectors dominate the total parameters in DeepFwFM, a $S_{\text{emb}}\%$ sparsity on the embedding vectors leads to the total memory reduction by roughly $1/(1-S_{\text{emb}}\%)$ times.

\section{Experiments}

\subsection{Experimental setup}
\subsubsection{Data sets}

1. Criteo Dataset: It is a well-known benchmark dataset for CTR prediction \cite{criteo}. It contains 45 million samples and each sample has 13 numerical features (counting) and 26 categorical features. We adopt the log transformation of $\log(x)^2$ if $x>2$ proposed by the winner of Criteo Competition \footnote{https://www.csie.ntu.edu.tw/~r01922136/kaggle-2014-criteo.pdf} to normalize the numerical features. We count the frequency of categorical features and treat all the features with a frequency less than 8 as unknown features. We randomly split the datasets into two parts: 90\% is used for training and the rest is left for testing. 2. Avazu Dataset: We use the 10 days of click-through log on users' mobile behaviors and randomly split 80\% of the samples for training and leave the rest for testing. We treat the features with frequency less than 5 as unknown and replace them by a field-wise default feature. A description of the two datasets is shown in Table.\ref{tab:table}. 

\begin{table}
 \caption{Statistics of datasets}
  \centering
  \begin{tabular}{lccccr}
    \toprule
    % \multicolumn{4}{c}{Criteo Dataset}                   \\
    % \midrule
    Data     & Training set  & \#  Fields  & \# Numerical   & \# Features & \\
    \midrule
    Criteo        &  41.3M & 39 & 13 & 1.33M    \\
    
    Avazu       & 32.3M & 23 & 0 & 1.54M \\
    \bottomrule
  \end{tabular}
  \label{tab:table}
%   \vspace{1em}
\end{table}

\subsubsection{Evaluation metrics}

To evaluate the prediction performance on Criteo dataset and Avazu dataset, we use LogLoss and AUC where LogLoss is the cross-entropy loss to evaluate the performance of a classification model and AUC is the area under the ROC curve to measure the probability that a random positive sample is ranked higher than a random negative sample.

\subsubsection{Baselines}

Among the popular embedding-based neural networks such as PNN \cite{PNN}, Deep \& Wide \cite{deepwide}, Deep Crossing \cite{deepcrossing}, Deep Cross \cite{deepcross}, AutoInt \cite{autoint}, DeepFM \cite{deepfm}, NFM \cite{NFM} and xDeepFM \cite{xdeepfm}, we choose the last three because they have similar architectures to DeepFwFM and they are also the state-of-the-art models for CTR prediction \footnote{We didn't compare DeepCross because it is a special case of xDeepFM.}. As a result, the 6 baseline models to evaluate DeepFwFM are LR (Logistic regression), FM \cite{FM}, FwFM \cite{fwfm}, DeepFM \cite{deepfm}, NFM \cite{NFM}, xDeepFM \cite{xdeepfm}.

\subsubsection{Implementation details}

We train our model using PyTorch. To make a fair comparison on Criteo dataset, we follow the parameter settings in \cite{deepfm, xdeepfm} and set the learning rate to $0.001$. The embedding size is set to 10. The default settings for the DNN components of DeepFM, NFM, and xDeepFM are: (1) the network architecture is $400\times 400\times 400$; (2) the dropout rate is 0.5. Specifically for xDeepFM, the cross layer in the CIN architecture is $100\times 100 \times 50$. We fine-tuned the $L_2$ penalty and set it to 3e-7. We use the Adam optimizer \cite{adam} for all of the experiments and the minibatch is chosen as 2048. Regarding Avazu dataset, we keep the same settings except that the embedding size is 20, the $L_2$ penalty is 6e-7, and the DNN network structure is $300\times 300\times 300$. Regarding the training time (not inference time) in practice, all the models don't differ each other too much. FwFM and DeepFwFM are slightly faster than DeepFM and xDeepFM due to the innovations in the linear terms, owing to the innovations in the inner products of FwFM and DeepFwFM.

\subsection{DeepFwFM v.s. other dense models}
\label{dense_model}

The evaluations of dense models without pruning show the maximum potential that the over-parameterized models perform. From Table \ref{criteo_data}, we observe that LR underperforms the other methods by at least 0.7\% on Criteo dataset and 1.7\% on Avazu dataset in terms of AUC, which shows that feature interactions are critical to improving the CTR prediction. Most of the embedding-based neural networks outperform the low-order methods such as LR and FM, implying the importance of modeling high-order feature interactions. However, the low-order FwFM still wins over NFM and DeepFM, showing the strength of field matrix $\bm{R}$ to learn second-order feature interactions to adapt to the local geometry. 

NFM utilizes a black-box DNN to implicitly learn the low-order and high-order feature interactions, which may potentially over-fit the datasets due to the lack of mechanism to identify the low-order feature interactions explicitly. Among all the embedding-based neural network models, \emph{xDeepFM and DeepFwFM achieves the best result on Criteo dataset and Avazu dataset} and outperform the other models by roughly 0.7\% on Criteo dataset and 0.4\% on Avazu dataset in terms of AUC. However, \emph{the inference time of xDeepFM is almost ten times longer than DeepFwFM on Criteo dataset}, showing the inefficiency in real-time predictions for large-scale ad serving systems.

\subsection{The sparse model: DeepLight}
\label{prune_exp}

Following Alg. \ref{prune} with damping ratios $\mathcal{D}=0.99$ and $\Omega=100$, we first train the network by 2 epochs for warm-ups, and then run 8 epochs for the pruning experiments. We prune the network every 10 iterations to reduce the computational cost.

\begin{table*}
 \caption{DeepFwFMs with sparse DNN components/ embedding vectorss v.s. DeepFwFMs with smaller DNN components/ embedding vectors. The DeepFwFM model with X nodes in each DNN layer is referred to as N-X; The DeepFwFM model with embedding size X is referred to as E-X. The baselines are chosen to have a close number of parameters of the sparse network.}
  \centering
  \small
  \begin{tabular}{llc|llc||llc|llc}
    \toprule
    \multicolumn{6}{c||}{Criteo} & \multicolumn{6}{c}{Avazu} \\
    \midrule
    \multirow{2}{*}{Sparse model}  & \multicolumn{2}{c|}{Test} & \multirow{2}{*}{Model} & \multicolumn{2}{c||}{Test} & \multirow{2}{*}{Sparse Model}  & \multicolumn{2}{c|}{Test} & \multirow{2}{*}{Model} & \multicolumn{2}{c}{Test} \\
     & Logloss     & AUC & &  Logloss     & AUC &  & LogLoss     & AUC & &  LogLoss     & AUC\\
    \midrule
    No Pruning   &  0.4403  & 0.8115 & N-400 & 0.4403 & 0.8115 &  No Pruning &   0.3751 & 0.7893   & N-300 &   0.3751 & 0.7893   \\
    \midrule
     D-90\% \& R-0\% \& F-0\%  &  \textbf{0.4398}  & \textbf{{0.8120}}  & N-87 & 0.4414 & 0.8104 & D-90\% \& R-0\% \& F-0\% &  \textbf{0.3747} & \textbf{0.7898}  & N-57 &   0.3757 & 0.7883  \\
     D-95\% \& R-0\% \& F-0\%  &  0.4399  & \textbf{{0.8120}}  & N-51 & 0.4421 & 0.8098 & D-95\% \& R-0\% \& F-0\% &  0.3747 & 0.7896  & N-32 &   0.3762 & 0.7875  \\
     D-98\% \& R-0\% \& F-0\% &  0.4401  & 0.8117  & N-25 &  0.4429 & 0.8089  & D-99\% \& R-0\% \& F-0\% &  0.3748 & 0.7895  & N-9 &  0.3769 & 0.7864   \\
     D-99\% \& R-0\% \& F-0\% &  0.4405  & 0.8113 & N-15 & 0.4438 & 0.8078 & D-99.5\% \& R-0\% \& F-0\% &  0.3749 & 0.7892 & N-6 &   0.3771 & 0.7862   \\
    \bottomrule
    \midrule
    No Pruning   &  0.4403  & 0.8115 & E-10 & 0.4403  & 0.8115 & No Pruning   &  0.3751 & 0.7893  & E-20 &  0.3751 & 0.7893   \\
    \midrule
     D-0\% \& R-0\% \& F-20\%  &   0.4402 & 0.8116   & E-8 & 0.4404 & 0.8115 &  D-0\% \& R-0\% \& F-20\% &  \textbf{0.3750} & \textbf{0.7895} & E-16 & 0.3752 & 0.7890    \\
     D-0\% \& R-0\% \& F-40\%  & \textbf{0.4401}  &  \textbf{{0.8118}} & E-6  & 0.4407 & 0.8113 & D-0\% \& R-0\% \& F-40\% & 0.3751 & 0.7891 & E-12 & 0.3750 & 0.7889  \\
     D-0\% \& R-0\% \& F-60\%  &  0.4404  & 0.8116  & E-4 & 0.4412 & 0.8106 &  D-0\% \& R-0\% \& F-60\% & 0.3762 & 0.7881 &  E-8 &   0.3765 & 0.7874  \\
     D-0\% \& R-0\% \& F-80\%  &  0.4406  & 0.8114  & E-2 & 0.4423 & 0.8094 &  D-0\% \& R-0\% \& F-80\% & 0.3773 & 0.7857 &  E-4 &   0.3770 & 0.7859  \\
    \bottomrule
  \end{tabular}
  \label{prune_deep}
  \vspace{1em}
\end{table*}

\subsubsection{\textbf{DNN pruning for accelerations}}
When we prune the DNN component, only the weights of the DNN component are pruned. The biases of DNN, the field matrix $\bm{R}$ and the parameters in the embedding layer is treated as usual. We try different pruning rates to study the prediction performance and deep model accelerations. To show the superiority of network pruning on a large network over training from a smaller one, we compare the networks with different sparse rates to the networks of smaller structures. 

As shown in Table.\ref{prune_deep} (top), we see the DeepFwFMs with sparse DNN components outperforms the dense DeepFwFMs even when the sparse rate is as high as 95\% on Criteo dataset. This phenomenon remains the same until we increase the sparsity to 99\% on Criteo dataset. By contrast, the corresponding small networks with similar number of parameters such as N-25 \footnote{A model with 25 nodes in each DNN layer is referred to as N-25.} and N-15 obtain much worse results than the original N-400, \emph{showing the power of pruning an over-parametrized network over training from a smaller one}. On Avazu dataset, we obtain the same conclusions. The sparse model obtains the best prediction performance with 90\% sparsity and only goes worse when the sparsity is larger than 99.5\%. 

\begin{figure}[h!]
\centering
\vspace{-0.1in}
  \subfigure[Criteo.]{\includegraphics[scale=0.33]{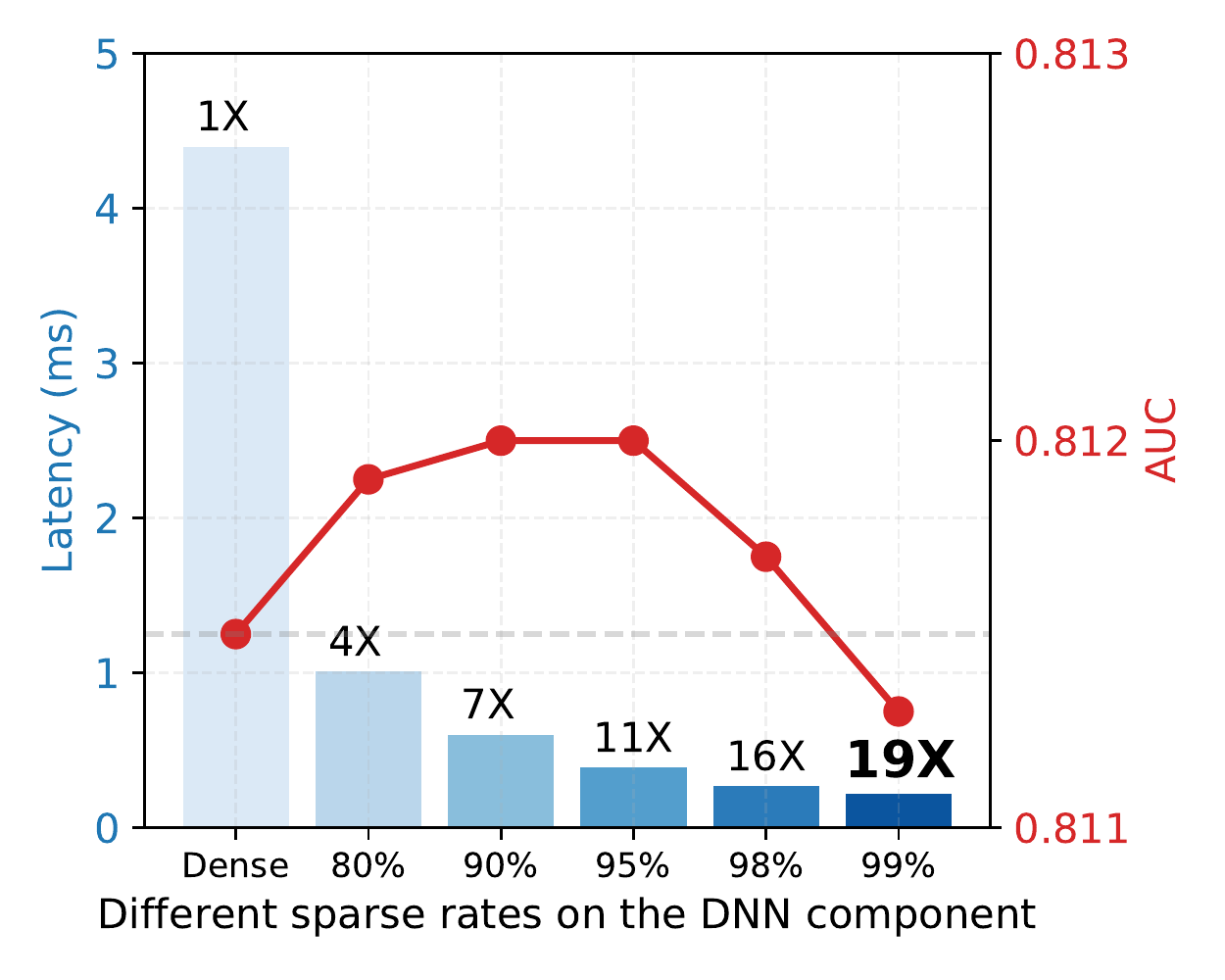}}
  \subfigure[Avazu.]{\includegraphics[scale=0.33]{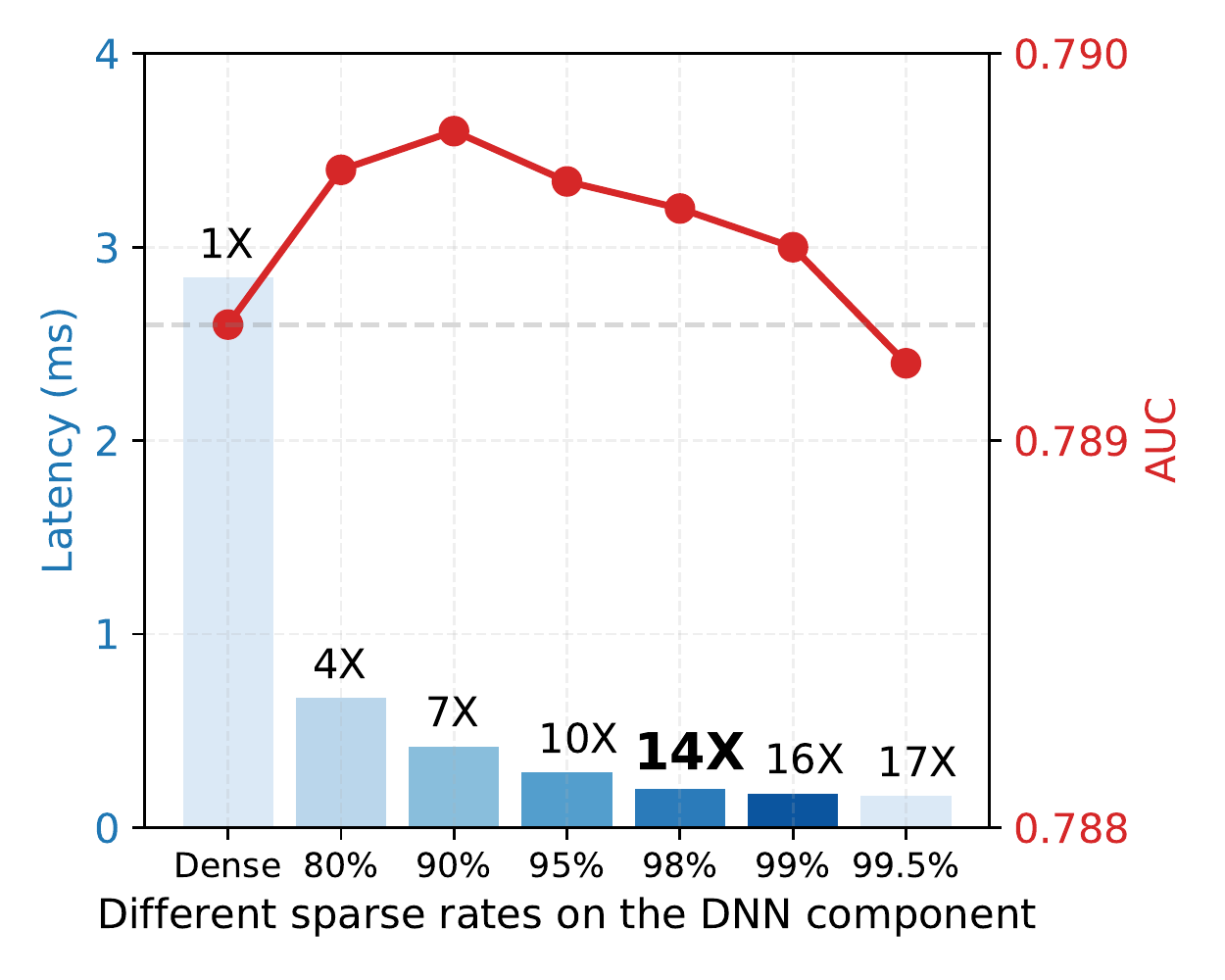}}
  \vspace{-0.1in}
  \caption{DNN pruning for accelerations.}
  \label{prune_fig_dnns}
%   \vspace{-0.5em}
\end{figure}

Regarding the deep model acceleration, we see from Fig.\ref{prune_fig_dnns} that a larger sparsity always brings a lower latency and when the sparsity is 98\% on Criteo dataset and 99\% on Avazu dataset, we realize the performance is still surprisingly better than the original dense network and we achieve as large as 16X speed-ups on both datasets.

\subsubsection{\textbf{Pruning of the field matrix }$\bm{R}$\textbf{ for accelerations}}

\begin{figure}[h!]
\centering
\vspace{-0.1in}
  \subfigure[Criteo]{\includegraphics[scale=0.33]{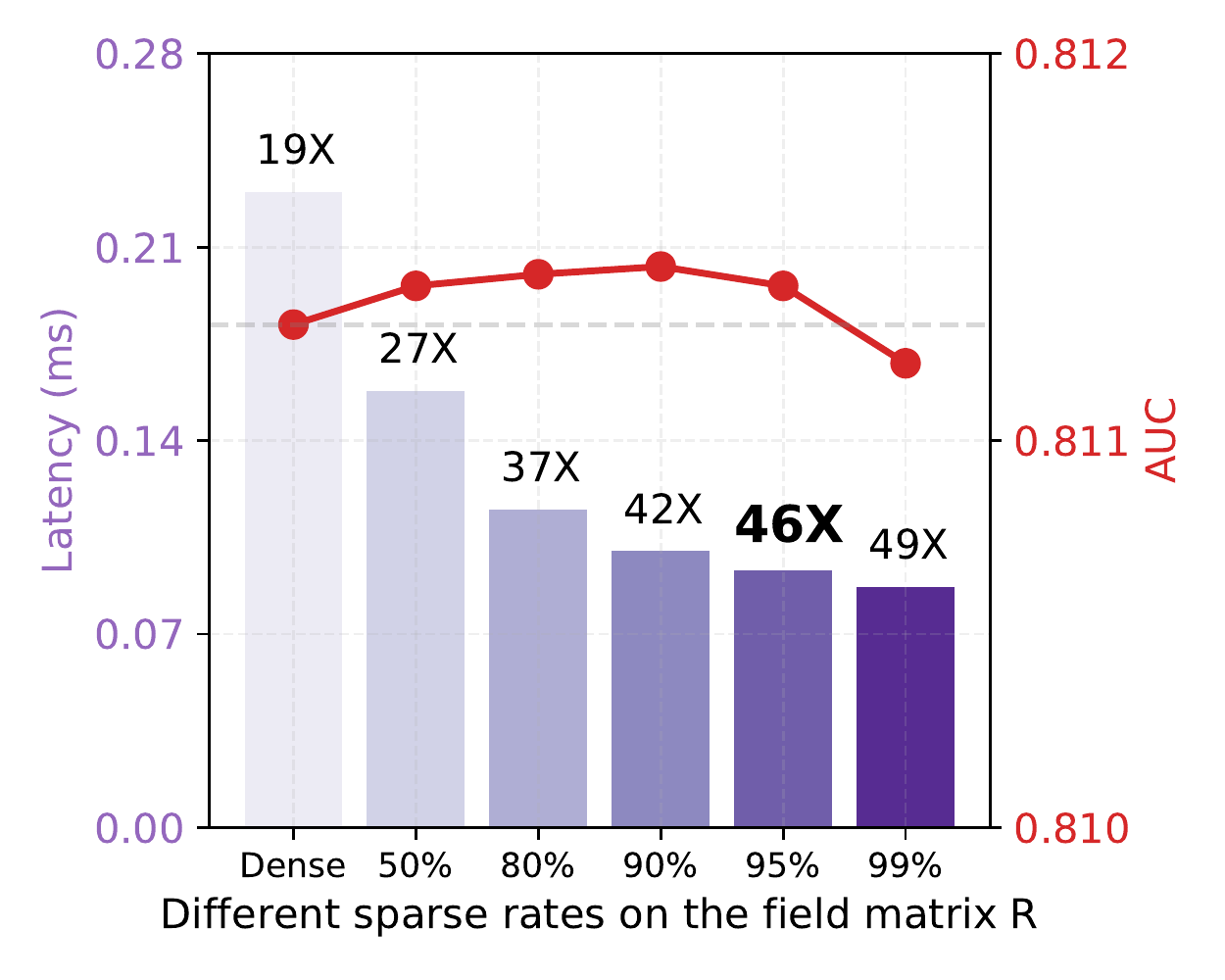}}
  \subfigure[Avazu]{\includegraphics[scale=0.33]{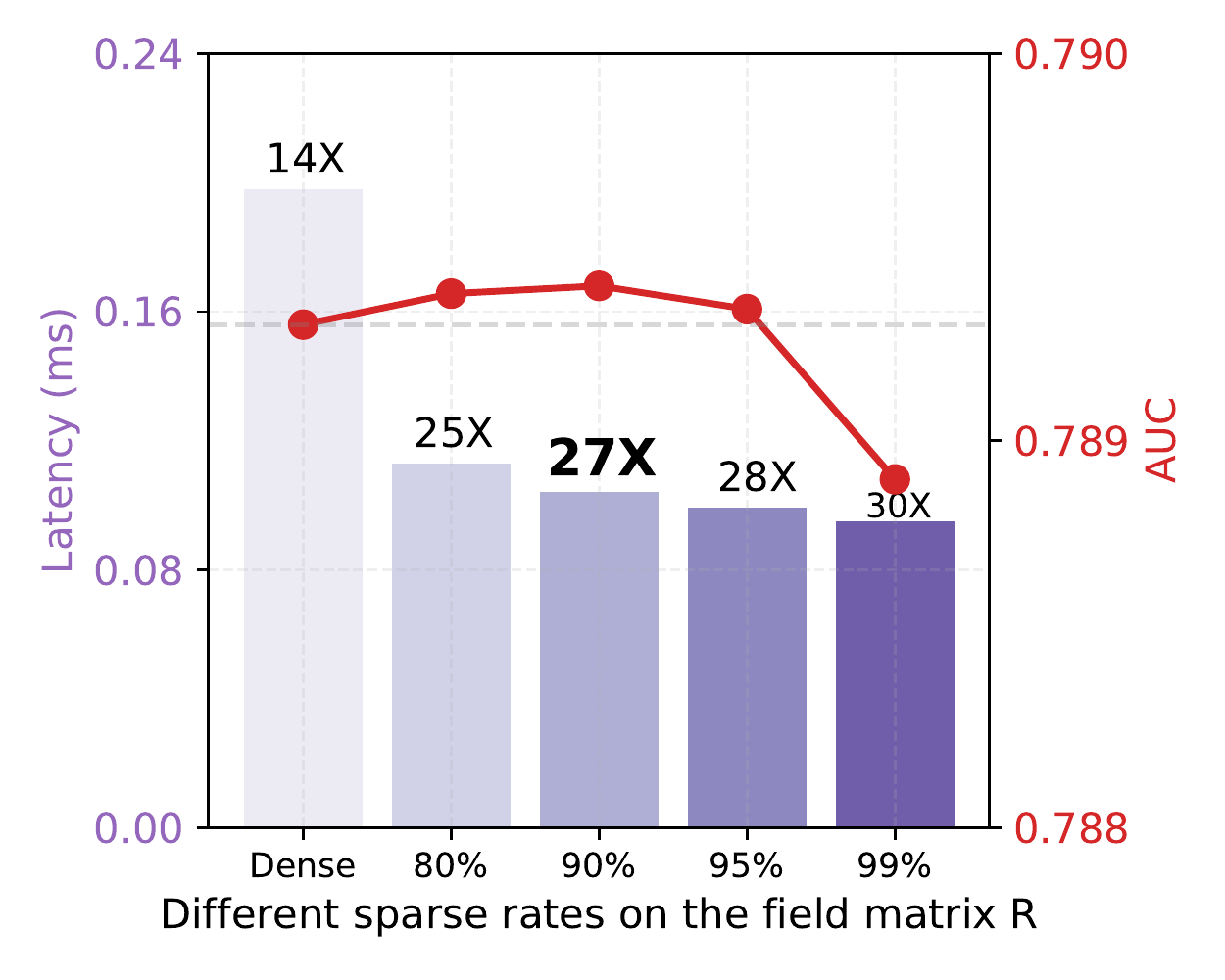}}
  \vspace{-0.05in}
  \caption{Pruning of the field matrix $\bm{R}$ for accelerations. }
   \label{prune_fig_Rs}
%   \vspace{-0.5em}
\end{figure}

After applying a high sparsity on the DNN component, we already obtain significant speed-ups which is close to 20X. \emph{To further boost the accelerations, increasing the sparsity on the DNN component may risk in decreasing the performance} and doesn't yield obvious accelerations due to the overhead in matrix computations. Recall from Fig.\ref{sparse_r} that the field matrix $\bm{R}$ possesses an approximately sparse structure. This motivates us to further prune the field matrix $\bm{R}$ to obtain deep accelerations. Given a 99\% sparsity for the DNN component on the Criteo dataset (98\% sparsity on the Avazu dataset), we study the sparse model performance based on different sparse rates in the field matrix $\bm{R}$. From Fig.\ref{prune_fig_Rs}, we observe that we can adopt up-to 95\% sparsity on the field matrix $\bm{R}$ without sacrificing the performance. Additionally, the predictions can be further accelerated by two to three times. As a result, we can eventually obtain 46X and 27X speed-ups without sacrificing the performance. 

%%%%%%

\subsubsection{\textbf{Pruning of embedding vectors for memory savings}}
As to the pruning of embeddings\footnote{To the best of our knowledge, this is the first structural pruning applied in embedding layers for memory savings and robust predictions by eliminating noisy estimates.}, we find that setting a global threshold for the embeddings of all fields obtains a slightly better performance than setting individual thresholds for the embedding vector from each field. Therefore, we conduct the experiments based on a global threshold. 

\begin{figure}[h!]
\centering
\vspace{-0.1in}
  \subfigure[Criteo]{\includegraphics[scale=0.33]{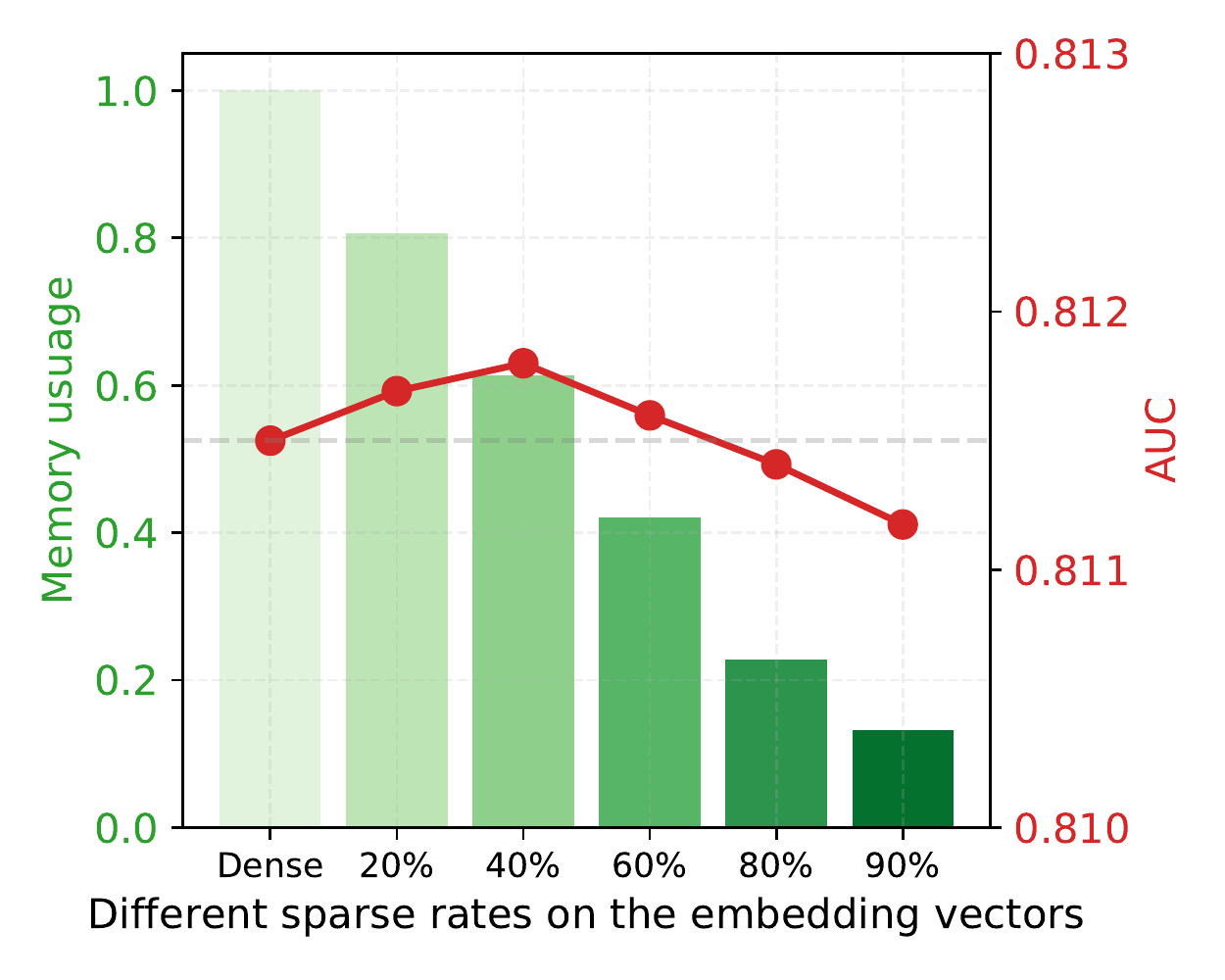}}\label{fig: fm2a} 
  \subfigure[Avazu]{\includegraphics[scale=0.33]{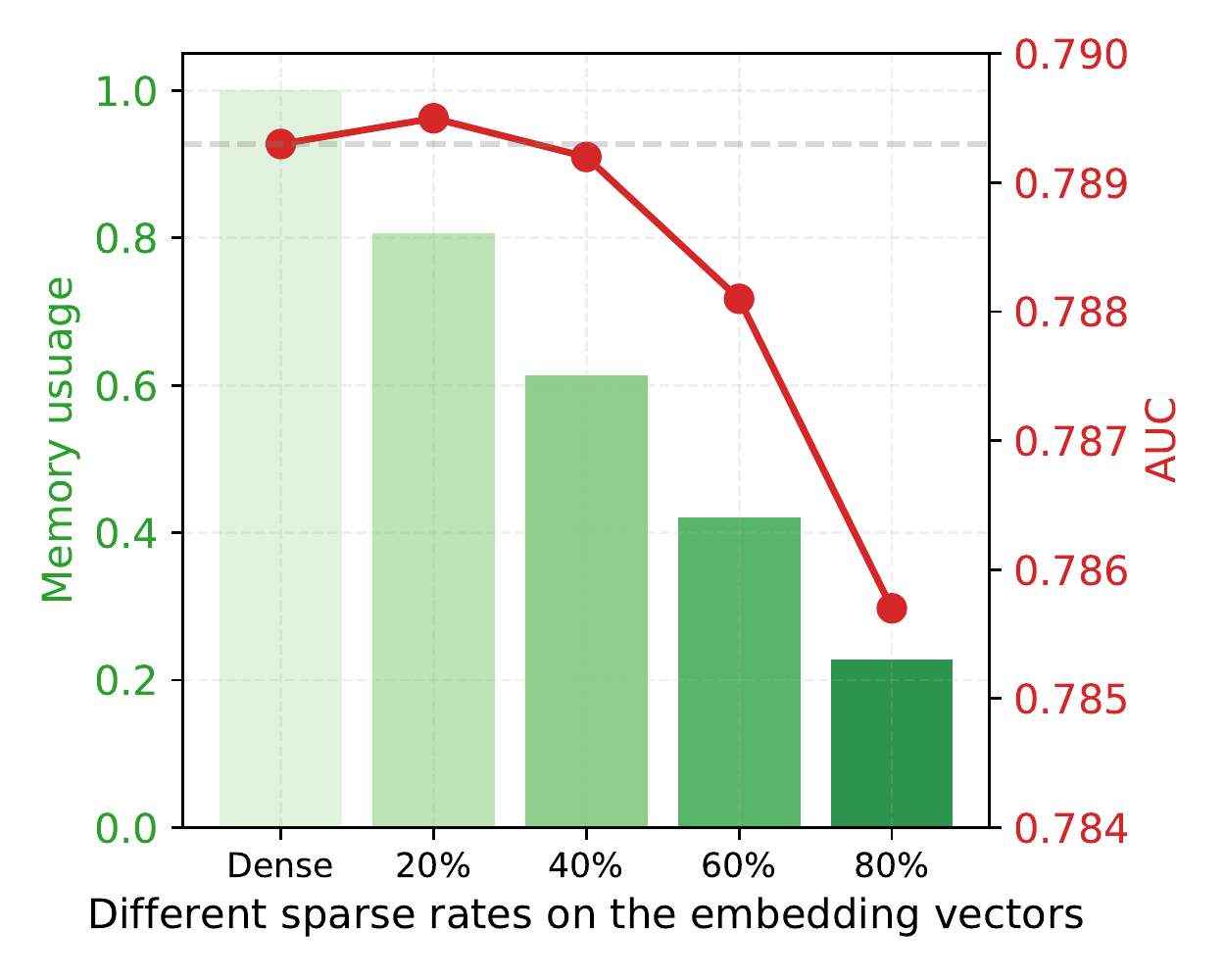}}\label{fig: avazu_2b}
   \vspace{-0.05in}
  \caption{Pruning of embedding vectors for memory savings.}
  \label{prune_fig_emb}
%   \vspace{1em}
\end{figure}

\begin{table*}
 \caption{Structural pruning of DeepFwFM on Criteo dataset. D-90\% \& R-90\% \& F-40\% is short for the sparse DeepFwFM which has 90\% sparse rate on the DNN component and the field matrix $\bm{R}$ and a 40\% sparse rate on the embedding vectors.}
  \centering
  \small
  \begin{tabular}{lllccccccrrc}
    \toprule
    \multirow{2}{*}{Dataset}& \multirow{2}{*}{Goal}& \multirow{2}{*}{Structural Pruning}  &  \multicolumn{2}{c}{Test} & \multirow{2}{*}{\# Parameters} & \multirow{2}{*}{Latency ($ms$)}  \\
    & &   & Logloss     & AUC & \\
    \midrule
    \multirow{4}{*}{Criteo} & None & No Pruning   &  0.4403  & 0.8116 & 13,739,321 & 4.271 \\
     & High performance & D-90\% \& R-90\% \& F-40\% &  \textbf{0.4395}  & \textbf{{0.8123}}  & 8,012,094 & 0.469 \\
     & Low memory & D-90\% \& R-90\% \& F-90\% &   0.4404  & 0.8114  & \textbf{1,376,431} & 0.472 \\
     & {Low latency} & D-99\% \& R-95\% \& F-40\%  &  0.4405  & 0.8114  &   7,413,578 & \textbf{0.093} \\
    \midrule
    \multirow{4}{*}{Avazu} & None & No Pruning   &   0.3751 & 0.7893   & 31,208,053  & 2.824 \\
    & High performance & D-90\% \& R-90\% \& F-20\%  & \textbf{0.3748}  & \textbf{0.7897}   & 24,808,262  & 0.422  \\
     & Low memory & D-90\% \& R-90\% \& F-60\%  & 0.3753  & 0.7892   &  \textbf{9,322,791} & 0.318  \\
     & Low latency & D-98\% \& R-90\% \& F-0\%  &    0.3753 & 0.7894    & 30,859,675   & \textbf{0.104}
      \\
     \bottomrule
  \end{tabular}
  \label{prune_all_criteo}
%   \vspace{-0.5em}
\end{table*}

As shown in Fig.\ref{prune_fig_emb}, Criteo can adopt a high sparse rate, such 80\%, to remain the same performance on Criteo dataset; by contrast, the model is sensitive on Avazu dataset and starts to decrease the performance when a 60\% sparsity is applied. From Table. \ref{prune_deep} (bottom), we see most of the models outperforms the baseline models (referred to as E-X) with a smaller embedding size, which sheds light on the use of large embedding sizes and pruning techniques to over-parameterize the network to avoid over-fitting.

\subsubsection{\textbf{Structural pruning of DeepFwFM}}
From the above experiments, we see that the DNN component and the field matrix $\bm{R}$ accept a much higher sparse rate to remain the same prediction performance, which inspires us to \emph{apply different pruning rates on the hybrid components}. 

As shown in Table.\ref{prune_all_criteo}, for the \emph{performance-driven tasks}, we can improve the state-of-the-art AUC from 0.8116 to 0.8223 on Criteo dataset via a sparse DeepFwFM where 90\% of the parameters in both the DNN component and the field matrix $\bm{R}$ and 40\% of the parameters in the embedding vectors are pruned, and such model is denoted by D-90\% \& R-90\% \& F-40\%. On Avazu dataset, a sparse DeepFwFM with structure D-90\% \& R-90\% \& F-20\% further improves the state-of-the-art AUC from 0.7893 to 0.7897. For the \emph{memory-driven tasks}, the memory savings are up to 10X and 2.5X on Criteo dataset and Avazu dataset, respectively. For the \emph{latency-driven tasks}, we achieve 46X speed-ups on Criteo dataset using a DeepLight with the structure D-99\% \& R-95\% \& F-40\% and 27X speed-ups on Avazu dataset using the structure D-98\% \& R-90\% \& F-0\% without loss of accuracy.

\subsection{{DeepLight v.s. other sparse models}}

For the other models, we also try the corresponding best structure for accelerating the predictions without sacrificing the performance. With respect to the CIN component in xDeepFM, we denote the 99\% sparsity on the CIN component by C-99\%. We report the results in Table.\ref{sparse_criteo_data} and observe that all the embedding based neural networks adopt high sparse rates to maintain the performance. Moreover, DeepLight is comparable to sparse DeepFM and sparse NFM in terms of prediction time but improves the AUC by at least 0.8\% on Criteo dataset and 0.4\% on Avazu dataset. \emph{DeepLight obtains a similar prediction performance as xDeepFM but is almost 10X faster}. This shows the superiority of DeepLight over sparse DeepFM, sparse NFM, sparse xDeepFM in large-scale online ad serving systems for both fast and accurate predictions.

\begin{table*}
 \caption{Evaluation of sparse models on Criteo and Avazu datasets. For each individual model, we only report the most efficient structure that yields the best accelerations with almost no sacrifice on the prediction performance. }
  \centering
  \small
  \begin{tabular}{lcccc|cccc}
    \toprule
    \multicolumn{5}{c|}{Criteo} & \multicolumn{4}{c}{Avazu} \\  
    \midrule
    \multirow{2}{*}{Models}    & \multicolumn{2}{c}{Test} & \multirow{2}{*}{Structure}  & \multirow{2}{*}{Latency ($ms$)} & \multicolumn{2}{c}{Test} & \multirow{2}{*}{Structure} & \multirow{2}{*}{Latency ($ms$)}  \\
      & LogLoss   & AUC &  & & LogLoss & AUC & \\
    \midrule
    Sparse DeepFM    & 0.4496 & 0.8032 &  D-98\% \& F-40\% & 0.114 & 0.3782  & 0.7851 & D-98\% \& F-20\% & \textbf{0.102}  \\
    Sparse NFM        &   0.4494 & 0.8031  & D-98\% \& F-40\% & 0.114  & 0.3778  & 0.7854 & D-98\% \& F-20\%  & \textbf{0.102}    \\
    Sparse xDeepFM     & 0.4421 & 0.8102 & \small{D-99\% \& C-99\% \& F-40\%}  &  0.907 & 0.3750  & 0.7893  &  \small{D-98\% \& C-98\% \& F-0\%} & 0.927   \\
    \midrule
    \textbf{DeepLight}    & \textbf{0.4405} & \textbf{0.8114}  & \small{D-99\% \& R-95\% \& F-40\%}  &  \textbf{0.093} & \textbf{0.3753} & \textbf{0.7894}  & \small{D-98\% \& R-90\% \& F-0\%} & 0.104  \\
    \bottomrule
  \end{tabular}
  \vspace{-0.5em}
  \label{sparse_criteo_data}
\end{table*}

\section{Conclusions}

In this paper, we propose the lightweight DeepLight model for efficient CTR predictions. A key advantage of this model is that the field matrix $\bm{R}$ not only provides a robust matrix decomposition to improve the performance with little costs, but also possesses a highly-sparse structure with acceleration potentials after pruning. To the best of our knowledge, this is the first work of network pruning applied to the area of CTR prediction in online advertising to solve the high-latency issues. We observe that network pruning is not only powerful to prune redundant parameters to alleviate over-fitting but also achieves significant acceleration on the inference time and shows a pronounced reduction on the memory usage with little impact on the prediction performance. %This strategy overcomes the challenge of efficient online advertising with the embedding-based neural networks. 

\section*{Acknowledgment} 
Lin acknowledges the support from NSF (DMS-1555072, DMS-1736364), BNL Subcontract 382247, W911NF-15-1-0562, and DE-SC0021142.

\bibliographystyle{ACM-Reference-Format}
\bibliography{sample-base}


\begin{thebibliography}{39}

%%% the Web address.

\ifx \showCODEN    \undefined \def \showCODEN     #1{\unskip}     \fi
\ifx \showDOI      \undefined \def \showDOI       #1{#1}\fi
\ifx \showISBNx    \undefined \def \showISBNx     #1{\unskip}     \fi
\ifx \showISBNxiii \undefined \def \showISBNxiii  #1{\unskip}     \fi
\ifx \showISSN     \undefined \def \showISSN      #1{\unskip}     \fi
\ifx \showLCCN     \undefined \def \showLCCN      #1{\unskip}     \fi
\ifx \shownote     \undefined \def \shownote      #1{#1}          \fi
\ifx \showarticletitle \undefined \def \showarticletitle #1{#1}   \fi
\ifx \showURL      \undefined \def \showURL       {\relax}        \fi
% invisible to TeX
\providecommand\bibfield[2]{#2}
\providecommand\bibinfo[2]{#2}
\providecommand\natexlab[1]{#1}
\providecommand\showeprint[2][]{arXiv:#2}

\bibitem[\protect\citeauthoryear{Blondel, Fujino, Ueda, and Ishihata}{Blondel
  et~al\mbox{.}}{2016}]%
        {NIPS2016_6144}
\bibfield{author}{\bibinfo{person}{Mathieu Blondel}, \bibinfo{person}{Akinori
  Fujino}, \bibinfo{person}{Naonori Ueda}, {and} \bibinfo{person}{Masakazu
  Ishihata}.} \bibinfo{year}{2016}\natexlab{}.
\newblock \showarticletitle{Higher-Order {F}actorization {M}achines}. In
  \bibinfo{booktitle}{\emph{NIPS'16}}.
\newblock


\bibitem[\protect\citeauthoryear{Bureau}{Bureau}{2020}]%
        {financial_report}
\bibfield{author}{\bibinfo{person}{Interactive~Advertising Bureau}.}
  \bibinfo{year}{2020}\natexlab{}.
\newblock \showarticletitle{IAB internet advertising revenue report}.
\newblock In \bibinfo{booktitle}{\emph{Iab Pwc}}. \bibinfo{pages}{1--32}.
\newblock


\bibitem[\protect\citeauthoryear{Chen, Zhan, Ci, and Lin}{Chen
  et~al\mbox{.}}{2020}]%
        {FLEN20}
\bibfield{author}{\bibinfo{person}{Wenqiang Chen}, \bibinfo{person}{Lizhang
  Zhan}, \bibinfo{person}{Yuanlong Ci}, {and} \bibinfo{person}{Chen Lin}.}
  \bibinfo{year}{2020}\natexlab{}.
\newblock \showarticletitle{FLEN: Leveraging Field for Scalable CTR
  Prediction}. In \bibinfo{booktitle}{\emph{KDD'20 DLP workshop}}.
\newblock


\bibitem[\protect\citeauthoryear{Cheng, Koc, Harmsen, Shaked, Chandra, Aradhye,
  Anderson, Corrado, Chai, Ispir, Anil, Haque, Hong, Jain, Liu, and Shah}{Cheng
  et~al\mbox{.}}{2016}]%
        {deepwide}
\bibfield{author}{\bibinfo{person}{Heng-Tze Cheng}, \bibinfo{person}{Levent
  Koc}, \bibinfo{person}{Jeremiah Harmsen}, \bibinfo{person}{Tal Shaked},
  \bibinfo{person}{Tushar Chandra}, \bibinfo{person}{Hrishi Aradhye},
  \bibinfo{person}{Glen Anderson}, \bibinfo{person}{Greg Corrado},
  \bibinfo{person}{Wei Chai}, \bibinfo{person}{Mustafa Ispir},
  \bibinfo{person}{Rohan Anil}, \bibinfo{person}{Zakaria Haque},
  \bibinfo{person}{Lichan Hong}, \bibinfo{person}{Vihan Jain},
  \bibinfo{person}{Xiaobing Liu}, {and} \bibinfo{person}{Hemal Shah}.}
  \bibinfo{year}{2016}\natexlab{}.
\newblock \showarticletitle{Wide \& Deep Learning for Recommender Systems}.
\newblock  (\bibinfo{year}{2016}).
\newblock
\showeprint[arxiv]{1606.07792}


\bibitem[\protect\citeauthoryear{Courbariaux and Bengio}{Courbariaux and
  Bengio}{2016}]%
        {weight_quant1}
\bibfield{author}{\bibinfo{person}{Matthieu Courbariaux} {and}
  \bibinfo{person}{Yoshua Bengio}.} \bibinfo{year}{2016}\natexlab{}.
\newblock \showarticletitle{BinaryNet: Training Deep Neural Networks with
  Weights and Activations Constrained to +1 or -1}.
\newblock \bibinfo{journal}{\emph{CoRR}}.
\newblock


\bibitem[\protect\citeauthoryear{Das and Kempe}{Das and Kempe}{2011}]%
        {Abhimanyu11}
\bibfield{author}{\bibinfo{person}{Abhimanyu Das} {and} \bibinfo{person}{David
  Kempe}.} \bibinfo{year}{2011}\natexlab{}.
\newblock \showarticletitle{Submodular meets Spectral: Greedy Algorithms for
  Subset Selection, Sparse Approximation and Dictionary Selection}. In
  \bibinfo{booktitle}{\emph{ICML'11}}.
\newblock


\bibitem[\protect\citeauthoryear{Deng, Zhang, Liang, and Lin}{Deng
  et~al\mbox{.}}{2019}]%
        {deng2019}
\bibfield{author}{\bibinfo{person}{Wei Deng}, \bibinfo{person}{Xiao Zhang},
  \bibinfo{person}{Faming Liang}, {and} \bibinfo{person}{Guang Lin}.}
  \bibinfo{year}{2019}\natexlab{}.
\newblock \showarticletitle{An {A}daptive {E}mpirical {B}ayesian {M}ethod for
  {S}parse {D}eep {L}earning}. In \bibinfo{booktitle}{\emph{NeurIPS'19}}.
\newblock


\bibitem[\protect\citeauthoryear{Frankle and Carbin}{Frankle and
  Carbin}{2019}]%
        {frankle2018lottery}
\bibfield{author}{\bibinfo{person}{Jonathan Frankle} {and}
  \bibinfo{person}{Michael Carbin}.} \bibinfo{year}{2019}\natexlab{}.
\newblock \showarticletitle{The lottery ticket hypothesis: Finding sparse,
  trainable neural networks}.
\newblock \bibinfo{journal}{\emph{ICLR'19}}.
\newblock


\bibitem[\protect\citeauthoryear{Graepel, Candela, Borchert, and
  Herbrich}{Graepel et~al\mbox{.}}{2010}]%
        {Graepel}
\bibfield{author}{\bibinfo{person}{Thore Graepel},
  \bibinfo{person}{Joaquin~Quiñonero Candela}, \bibinfo{person}{Thomas
  Borchert}, {and} \bibinfo{person}{Ralf Herbrich}.}
  \bibinfo{year}{2010}\natexlab{}.
\newblock \showarticletitle{Web-scale Bayesian Click-through Rate Prediction
  for Sponsored Search Advertising in {M}icrosoft’s {B}ing Search Engine}. In
  \bibinfo{booktitle}{\emph{ICML’10}}.
\newblock


\bibitem[\protect\citeauthoryear{Guo, Tang, Ye, Li, and He}{Guo
  et~al\mbox{.}}{2017}]%
        {deepfm}
\bibfield{author}{\bibinfo{person}{Huifeng Guo}, \bibinfo{person}{Ruiming
  Tang}, \bibinfo{person}{Yunming Ye}, \bibinfo{person}{Zhenguo Li}, {and}
  \bibinfo{person}{Xiuqiang He}.} \bibinfo{year}{2017}\natexlab{}.
\newblock \showarticletitle{DeepFM: {A} {F}actorization-{M}achine based
  {N}eural {N}etwork for {CTR} {P}rediction}.
\newblock In \bibinfo{booktitle}{\emph{IJCAI-17}}.
\newblock


\bibitem[\protect\citeauthoryear{Han, Mao, and Dally}{Han
  et~al\mbox{.}}{2016}]%
        {hansong16}
\bibfield{author}{\bibinfo{person}{Song Han}, \bibinfo{person}{Huizi Mao},
  {and} \bibinfo{person}{William~J. Dally}.} \bibinfo{year}{2016}\natexlab{}.
\newblock \showarticletitle{Deep Compression: Compressing Deep Neural Networks
  with Pruning, Trained Quantization and Huffman Coding}.
\newblock In \bibinfo{booktitle}{\emph{ICLR'16}}.
\newblock


\bibitem[\protect\citeauthoryear{Han, Pool, Tran, and Dally}{Han
  et~al\mbox{.}}{2015}]%
        {han2015learning}
\bibfield{author}{\bibinfo{person}{Song Han}, \bibinfo{person}{Jeff Pool},
  \bibinfo{person}{John Tran}, {and} \bibinfo{person}{William Dally}.}
  \bibinfo{year}{2015}\natexlab{}.
\newblock \showarticletitle{Learning both weights and connections for efficient
  neural network}. In \bibinfo{booktitle}{\emph{NIPS'15}}.
\newblock


\bibitem[\protect\citeauthoryear{He and Chua}{He and Chua}{2017}]%
        {NFM}
\bibfield{author}{\bibinfo{person}{Xiangnan He} {and}
  \bibinfo{person}{Tat{-}Seng Chua}.} \bibinfo{year}{2017}\natexlab{}.
\newblock \showarticletitle{Neural Factorization Machines for Sparse Predictive
  Analytics}. In \bibinfo{booktitle}{\emph{SIGIR'17}}.
\newblock


\bibitem[\protect\citeauthoryear{He, Pan, Jin, Xu, Liu, Xu, Shi, Atallah,
  Herbrich, Bowers, and Candela}{He et~al\mbox{.}}{2014}]%
        {Xinran_He}
\bibfield{author}{\bibinfo{person}{Xinran He}, \bibinfo{person}{Junfeng Pan},
  \bibinfo{person}{Ou Jin}, \bibinfo{person}{Tianbing Xu}, \bibinfo{person}{Bo
  Liu}, \bibinfo{person}{Tao Xu}, \bibinfo{person}{Yanxin Shi},
  \bibinfo{person}{Antoine Atallah}, \bibinfo{person}{Ralf Herbrich},
  \bibinfo{person}{Stuart Bowers}, {and} \bibinfo{person}{Joaquin~Quinonero
  Candela}.} \bibinfo{year}{2014}\natexlab{}.
\newblock \showarticletitle{Practical Lessons from Predicting Clicks on Ads at
  Facebook}. In \bibinfo{booktitle}{\emph{ADKDD'14}}.
\newblock


\bibitem[\protect\citeauthoryear{Hinton, Vinyals, and Dean}{Hinton
  et~al\mbox{.}}{2015}]%
        {distill}
\bibfield{author}{\bibinfo{person}{Geoffrey Hinton}, \bibinfo{person}{Oriol
  Vinyals}, {and} \bibinfo{person}{Jeffrey Dean}.}
  \bibinfo{year}{2015}\natexlab{}.
\newblock \showarticletitle{Distilling the Knowledge in a Neural Network}. In
  \bibinfo{booktitle}{\emph{NIPS Deep Learning and RL Workshop}}.
\newblock


\bibitem[\protect\citeauthoryear{Juan, Zhuang, Chin, and Lin}{Juan
  et~al\mbox{.}}{2016}]%
        {ffm}
\bibfield{author}{\bibinfo{person}{Yuchin Juan}, \bibinfo{person}{Yong Zhuang},
  \bibinfo{person}{Wei-Sheng Chin}, {and} \bibinfo{person}{Chih-Jen Lin}.}
  \bibinfo{year}{2016}\natexlab{}.
\newblock \showarticletitle{Field-aware Factorization Machines for CTR
  Prediction}. In \bibinfo{booktitle}{\emph{RecSys'16}}.
\newblock


\bibitem[\protect\citeauthoryear{Kingma and Ba}{Kingma and Ba}{2015}]%
        {adam}
\bibfield{author}{\bibinfo{person}{Diederik~P. Kingma} {and}
  \bibinfo{person}{Jimmy Ba}.} \bibinfo{year}{2015}\natexlab{}.
\newblock \showarticletitle{Adam: {A} Method for {S}tochastic {O}ptimization}.
\newblock In \bibinfo{booktitle}{\emph{ICLR'15}}.
\newblock


\bibitem[\protect\citeauthoryear{Labs}{Labs}{2014}]%
        {criteo}
\bibfield{author}{\bibinfo{person}{Criteo Labs}.}
  \bibinfo{year}{2014}\natexlab{}.
\newblock \showarticletitle{Display Advertising Challenge}.
\newblock In \bibinfo{booktitle}{\emph{Kaggle}}.
\newblock


\bibitem[\protect\citeauthoryear{Li, Chen, Carlson, and Carin}{Li
  et~al\mbox{.}}{2016}]%
        {Li16}
\bibfield{author}{\bibinfo{person}{Chunyuan Li}, \bibinfo{person}{Changyou
  Chen}, \bibinfo{person}{David Carlson}, {and} \bibinfo{person}{Lawrence
  Carin}.} \bibinfo{year}{2016}\natexlab{}.
\newblock \showarticletitle{Preconditioned {S}tochastic {G}radient {L}angevin
  {D}ynamics for {D}eep {N}eural {N}etworks}. In
  \bibinfo{booktitle}{\emph{AAAI'16}}.
\newblock


\bibitem[\protect\citeauthoryear{Li, Wang, and Kong}{Li et~al\mbox{.}}{2018}]%
        {deguang}
\bibfield{author}{\bibinfo{person}{Dawei Li}, \bibinfo{person}{Xiaolong Wang},
  {and} \bibinfo{person}{Deguang Kong}.} \bibinfo{year}{2018}\natexlab{}.
\newblock \showarticletitle{DeepRebirth: Accelerating Deep Neural Network
  Execution on Mobile Devices}. In \bibinfo{booktitle}{\emph{AAAI'18}}.
\newblock


\bibitem[\protect\citeauthoryear{Li, Kadav, Durdanovic, Samet, and Graf}{Li
  et~al\mbox{.}}{2017}]%
        {li17}
\bibfield{author}{\bibinfo{person}{Hao Li}, \bibinfo{person}{Asim Kadav},
  \bibinfo{person}{Igor Durdanovic}, \bibinfo{person}{Hanan Samet}, {and}
  \bibinfo{person}{Hans~Peter Graf}.} \bibinfo{year}{2017}\natexlab{}.
\newblock \showarticletitle{Pruning Filters for Efficient ConvNets}.
\newblock In \bibinfo{booktitle}{\emph{ICLR'17}}.
\newblock


\bibitem[\protect\citeauthoryear{Lian, Zhou, Zhang, Chen, Xie, and Sun}{Lian
  et~al\mbox{.}}{2018}]%
        {xdeepfm}
\bibfield{author}{\bibinfo{person}{Jianxun Lian}, \bibinfo{person}{Xiaohuan
  Zhou}, \bibinfo{person}{Fuzheng Zhang}, \bibinfo{person}{Zhongxia Chen},
  \bibinfo{person}{Xing Xie}, {and} \bibinfo{person}{Guangzhong Sun}.}
  \bibinfo{year}{2018}\natexlab{}.
\newblock \showarticletitle{xDeepFM: Combining Explicit and Implicit Feature
  Interactions for Recommender Systems}.
\newblock \bibinfo{journal}{\emph{KDD'18}}.
\newblock


\bibitem[\protect\citeauthoryear{Liu, Zhu, Li, Zhang, Lai, Tang, He, Li, and
  Yu}{Liu et~al\mbox{.}}{2020}]%
        {autofis}
\bibfield{author}{\bibinfo{person}{Bin Liu}, \bibinfo{person}{Chenxu Zhu},
  \bibinfo{person}{Guilin Li}, \bibinfo{person}{Weinan Zhang},
  \bibinfo{person}{Jincai Lai}, \bibinfo{person}{Ruiming Tang},
  \bibinfo{person}{Xiuqiang He}, \bibinfo{person}{Zhenguo Li}, {and}
  \bibinfo{person}{Yong Yu}.} \bibinfo{year}{2020}\natexlab{}.
\newblock \showarticletitle{AutoFIS: Automatic Feature Interaction Selection in
  Factorization Models for Click-Through Rate Prediction}.
\newblock In \bibinfo{booktitle}{\emph{KDD'20}}.
\newblock


\bibitem[\protect\citeauthoryear{Luo, Wang, Zhou, Yao, Tu, Chen, Yang, and
  Dai}{Luo et~al\mbox{.}}{2019}]%
        {autocross}
\bibfield{author}{\bibinfo{person}{Yuanfei Luo}, \bibinfo{person}{Mengshuo
  Wang}, \bibinfo{person}{Hao Zhou}, \bibinfo{person}{Quanming Yao},
  \bibinfo{person}{Wei{-}Wei Tu}, \bibinfo{person}{Yuqiang Chen},
  \bibinfo{person}{Qiang Yang}, {and} \bibinfo{person}{Wenyuan Dai}.}
  \bibinfo{year}{2019}\natexlab{}.
\newblock \showarticletitle{AutoCross: Automatic Feature Crossing for Tabular
  Data in Real-World Applications}.
\newblock In \bibinfo{booktitle}{\emph{KDD'19}}.
\newblock


\bibitem[\protect\citeauthoryear{Mackey, Talwalkar, and Jordan}{Mackey
  et~al\mbox{.}}{2015}]%
        {Michael_Jordan15}
\bibfield{author}{\bibinfo{person}{Lester Mackey}, \bibinfo{person}{Ameet
  Talwalkar}, {and} \bibinfo{person}{Michael~I. Jordan}.}
  \bibinfo{year}{2015}\natexlab{}.
\newblock \showarticletitle{Distributed {M}atrix {C}ompletion and {R}obust
  {F}actorization}.
\newblock \bibinfo{journal}{\emph{JMLR'15}}  \bibinfo{volume}{16}
  (\bibinfo{year}{2015}), \bibinfo{pages}{913--960}.
\newblock


\bibitem[\protect\citeauthoryear{McMahan, Holt, Sculley, Young, Ebner, Grady,
  Nie, Phillips, Davydov, Golovin, Chikkerur, Liu, Wattenberg, Hrafnkelsson,
  Boulos, and Kubica}{McMahan et~al\mbox{.}}{2013}]%
        {linear1}
\bibfield{author}{\bibinfo{person}{H.~Brendan McMahan}, \bibinfo{person}{Gary
  Holt}, \bibinfo{person}{D. Sculley}, \bibinfo{person}{Michael Young},
  \bibinfo{person}{Dietmar Ebner}, \bibinfo{person}{Julian Grady},
  \bibinfo{person}{Lan Nie}, \bibinfo{person}{Todd Phillips},
  \bibinfo{person}{Eugene Davydov}, \bibinfo{person}{Daniel Golovin},
  \bibinfo{person}{Sharat Chikkerur}, \bibinfo{person}{Dan Liu},
  \bibinfo{person}{Martin Wattenberg}, \bibinfo{person}{Arnar~Mar
  Hrafnkelsson}, \bibinfo{person}{Tom Boulos}, {and} \bibinfo{person}{Jeremy
  Kubica}.} \bibinfo{year}{2013}\natexlab{}.
\newblock \showarticletitle{Ad Click Prediction: a View from the Trenches}. In
  \bibinfo{booktitle}{\emph{KDD'13}}.
\newblock


\bibitem[\protect\citeauthoryear{Pan, Mao, Ruiz, Sun, and Flores}{Pan
  et~al\mbox{.}}{2019}]%
        {mt-fwfm}
\bibfield{author}{\bibinfo{person}{Junwei Pan}, \bibinfo{person}{Yizhi Mao},
  \bibinfo{person}{Alfonso~Lobos Ruiz}, \bibinfo{person}{Yu Sun}, {and}
  \bibinfo{person}{Aaron Flores}.} \bibinfo{year}{2019}\natexlab{}.
\newblock \showarticletitle{Predicting different types of conversions with
  multi-task learning in online advertising}. In
  \bibinfo{booktitle}{\emph{KDD'19}}.
\newblock


\bibitem[\protect\citeauthoryear{Pan, Xu, Ruiz, Zhao, Pan, Sun, and Lu}{Pan
  et~al\mbox{.}}{2018}]%
        {fwfm}
\bibfield{author}{\bibinfo{person}{Junwei Pan}, \bibinfo{person}{Jian Xu},
  \bibinfo{person}{Alfonso~Lobos Ruiz}, \bibinfo{person}{Wenliang Zhao},
  \bibinfo{person}{Shengjun Pan}, \bibinfo{person}{Yu Sun}, {and}
  \bibinfo{person}{Quan Lu}.} \bibinfo{year}{2018}\natexlab{}.
\newblock \showarticletitle{Field-weighted Factorization Machines for
  Click-Through Rate Prediction in Display Advertising}. In
  \bibinfo{booktitle}{\emph{WWW'18}}.
\newblock


\bibitem[\protect\citeauthoryear{Qu, Cai, Ren, Zhang, Yu, Wen, and Wang}{Qu
  et~al\mbox{.}}{2016}]%
        {PNN}
\bibfield{author}{\bibinfo{person}{Yanru Qu}, \bibinfo{person}{Han Cai},
  \bibinfo{person}{Kan Ren}, \bibinfo{person}{Weinan Zhang},
  \bibinfo{person}{Yong Yu}, \bibinfo{person}{Ying Wen}, {and}
  \bibinfo{person}{Jun Wang}.} \bibinfo{year}{2016}\natexlab{}.
\newblock \showarticletitle{Product-based Neural Networks for User Response
  Prediction}.
\newblock \bibinfo{journal}{\emph{CoRR}} (\bibinfo{year}{2016}).
\newblock


\bibitem[\protect\citeauthoryear{Rendle}{Rendle}{2010}]%
        {FM}
\bibfield{author}{\bibinfo{person}{Steffen Rendle}.}
  \bibinfo{year}{2010}\natexlab{}.
\newblock \showarticletitle{Factorization Machines}. In
  \bibinfo{booktitle}{\emph{ICDM'10}}.
\newblock


\bibitem[\protect\citeauthoryear{Shan, Hoens, Jiao, Wang, Yu, and Mao}{Shan
  et~al\mbox{.}}{2016}]%
        {deepcrossing}
\bibfield{author}{\bibinfo{person}{Ying Shan}, \bibinfo{person}{T.~Ryan Hoens},
  \bibinfo{person}{Jian Jiao}, \bibinfo{person}{Haijing Wang},
  \bibinfo{person}{Dong Yu}, {and} \bibinfo{person}{JC Mao}.}
  \bibinfo{year}{2016}\natexlab{}.
\newblock \showarticletitle{Deep Crossing: Web-Scale Modeling Without Manually
  Crafted Combinatorial Features}. In \bibinfo{booktitle}{\emph{KDD'16}}.
\newblock


\bibitem[\protect\citeauthoryear{Shaparenko, Cetin, and Iyer}{Shaparenko
  et~al\mbox{.}}{2009}]%
        {Shaparenko}
\bibfield{author}{\bibinfo{person}{Benyah Shaparenko}, \bibinfo{person}{Ozgur
  Cetin}, {and} \bibinfo{person}{Rukmini Iyer}.}
  \bibinfo{year}{2009}\natexlab{}.
\newblock \showarticletitle{Data-driven Text Features for Sponsored Search
  Click Prediction}. In \bibinfo{booktitle}{\emph{ADKDD'09}}.
\newblock


\bibitem[\protect\citeauthoryear{Song, Shi, Xiao, Duan, Xu, Zhang, and
  Tang}{Song et~al\mbox{.}}{2019}]%
        {autoint}
\bibfield{author}{\bibinfo{person}{Weiping Song}, \bibinfo{person}{Chence Shi},
  \bibinfo{person}{Zhiping Xiao}, \bibinfo{person}{Zhijian Duan},
  \bibinfo{person}{Yewen Xu}, \bibinfo{person}{Ming Zhang}, {and}
  \bibinfo{person}{Jian Tang}.} \bibinfo{year}{2019}\natexlab{}.
\newblock \showarticletitle{AutoInt: Automatic Feature Interaction Learning via
  Self-Attentive Neural Networks}.
\newblock \bibinfo{journal}{\emph{CIKM'19}}.
\newblock


\bibitem[\protect\citeauthoryear{Wang, Fu, Fu, and Wang}{Wang
  et~al\mbox{.}}{2017}]%
        {deepcross}
\bibfield{author}{\bibinfo{person}{Ruoxi Wang}, \bibinfo{person}{Bin Fu},
  \bibinfo{person}{Gang Fu}, {and} \bibinfo{person}{Mingliang Wang}.}
  \bibinfo{year}{2017}\natexlab{}.
\newblock \showarticletitle{Deep {\&} Cross Network for Ad Click Predictions}.
\newblock
\showeprint[arxiv]{1708.05123}


\bibitem[\protect\citeauthoryear{Wang, Bian, Liu, Zhang, and Liu}{Wang
  et~al\mbox{.}}{2013}]%
        {Taifeng}
\bibfield{author}{\bibinfo{person}{Taifeng Wang}, \bibinfo{person}{Jiang Bian},
  \bibinfo{person}{Shusen Liu}, \bibinfo{person}{Yuyu Zhang}, {and}
  \bibinfo{person}{Tie-Yan Liu}.} \bibinfo{year}{2013}\natexlab{}.
\newblock \showarticletitle{Psychological Advertising: Exploring User
  Psychology for Click Prediction in Sponsored Search}.
\newblock \bibinfo{journal}{\emph{KDD'13}}.
\newblock


\bibitem[\protect\citeauthoryear{Wen, Wu, Wang, Chen, and Li}{Wen
  et~al\mbox{.}}{2016}]%
        {strucprunining}
\bibfield{author}{\bibinfo{person}{Wei Wen}, \bibinfo{person}{Chunpeng Wu},
  \bibinfo{person}{Yandan Wang}, \bibinfo{person}{Yiran Chen}, {and}
  \bibinfo{person}{Hai Li}.} \bibinfo{year}{2016}\natexlab{}.
\newblock \showarticletitle{Learning {S}tructured {S}parsity in {D}eep {N}eural
  {N}etworks}.
\newblock In \bibinfo{booktitle}{\emph{NIPS'16}}.
\newblock


\bibitem[\protect\citeauthoryear{Ye, Gong, Nie, Zhou, Klivans, and Liu}{Ye
  et~al\mbox{.}}{2020a}]%
        {ye_icml}
\bibfield{author}{\bibinfo{person}{Mao Ye}, \bibinfo{person}{Chengyue Gong},
  \bibinfo{person}{Lizhen Nie}, \bibinfo{person}{Denny Zhou},
  \bibinfo{person}{Adam Klivans}, {and} \bibinfo{person}{Qiang Liu}.}
  \bibinfo{year}{2020}\natexlab{a}.
\newblock \showarticletitle{Good Subnetworks Provably Exist: Pruning via Greedy
  Forward Selection}. In \bibinfo{booktitle}{\emph{ICML'20}}.
\newblock


\bibitem[\protect\citeauthoryear{Ye, Wu, and Liu}{Ye et~al\mbox{.}}{2020b}]%
        {ye_nips}
\bibfield{author}{\bibinfo{person}{Mao Ye}, \bibinfo{person}{Lemeng Wu}, {and}
  \bibinfo{person}{Qiang Liu}.} \bibinfo{year}{2020}\natexlab{b}.
\newblock \showarticletitle{Greedy Optimization Provably Wins the Lottery:
  Logarithmic Number of Winning Tickets is Enough}. In
  \bibinfo{booktitle}{\emph{NeurIPS'20}}.
\newblock


\bibitem[\protect\citeauthoryear{Zhou, Zhu, Song, Fan, Zhu, Ma, Yan, Jin, Li,
  and Gai}{Zhou et~al\mbox{.}}{2014}]%
        {DIN}
\bibfield{author}{\bibinfo{person}{Guorui Zhou}, \bibinfo{person}{Xiaoqiang
  Zhu}, \bibinfo{person}{Chengru Song}, \bibinfo{person}{Ying Fan},
  \bibinfo{person}{Han Zhu}, \bibinfo{person}{Xiao Ma},
  \bibinfo{person}{Yanghui Yan}, \bibinfo{person}{Junqi Jin},
  \bibinfo{person}{Han Li}, {and} \bibinfo{person}{Kun Gai}.}
  \bibinfo{year}{2014}\natexlab{}.
\newblock \showarticletitle{Deep Interest Network for Click-Through Rate
  Prediction}.
\newblock In \bibinfo{booktitle}{\emph{KDD'14}}.
\newblock


\end{thebibliography}

\end{document}